%% file: main.tex
\documentclass[10pt,twocolumn,letterpaper]{article}

\usepackage{cvpr}              


\definecolor{cvprblue}{rgb}{0.21,0.49,0.74}
\usepackage[pagebackref,breaklinks,colorlinks,allcolors=cvprblue]{hyperref}
\usepackage{float}
\usepackage{graphicx}
\usepackage{multirow}
\usepackage[normalem]{ulem}
\useunder{\uline}{\ul}{}
\usepackage{enumitem}
\usepackage{placeins}
\usepackage{caption}
\usepackage{capt-of}  
\usepackage{cuted}    
\usepackage{booktabs}       
\usepackage{amsfonts}       
\usepackage{nicefrac}       
\usepackage{microtype}      
\usepackage{xcolor}         

\newcommand{\IGNORE}[1]{}    
\IGNORE{
\typeout{freefloats...}                  
\setcounter{topnumber}{10}               
\setcounter{bottomnumber}{10}            
\setcounter{totalnumber}{10}             
\setcounter{dbltopnumber}{10}            
}

\def\paperID{5071} 
\def\confName{CVPR}
\def\confYear{2025}

\title{NSD-Imagery: A benchmark dataset for extending fMRI vision decoding methods to mental imagery}

\author{%
      Reese Kneeland\textsuperscript{1 2}\quad
      Paul S. Scotti\textsuperscript{3 4}\quad
      Ghislain St-Yves\textsuperscript{1}\quad
      Jesse Breedlove\textsuperscript{1}\and
      Kendrick Kay\textsuperscript{1}\quad
      Thomas Naselaris\textsuperscript{1} \\
      \vspace{0.2em}
    {\small\textsuperscript{1}University of Minnesota} \\
    {\small\textsuperscript{2}Alljoined} \\
    {\small\textsuperscript{3}Princeton Neuroscience Institute} \\
    {\small\textsuperscript{4}Formerly Stability AI/Medical AI Research Center (MedARC)} \\
    {\small Correspondence to: {\tt\small nase0005@umn.edu}} \\
}

\begin{document}
\maketitle
\input{body}
{
    \small
    \bibliographystyle{ieeenat_fullname}
    \bibliography{main}
}
\clearpage
\input{appendix}

\end{document}

%% file: body.tex
\begin{strip}\centering
\vspace{-45pt}       
\setlength{\abovecaptionskip}{2pt}    
\setlength{\belowcaptionskip}{-14pt}    
\includegraphics[width=\textwidth]{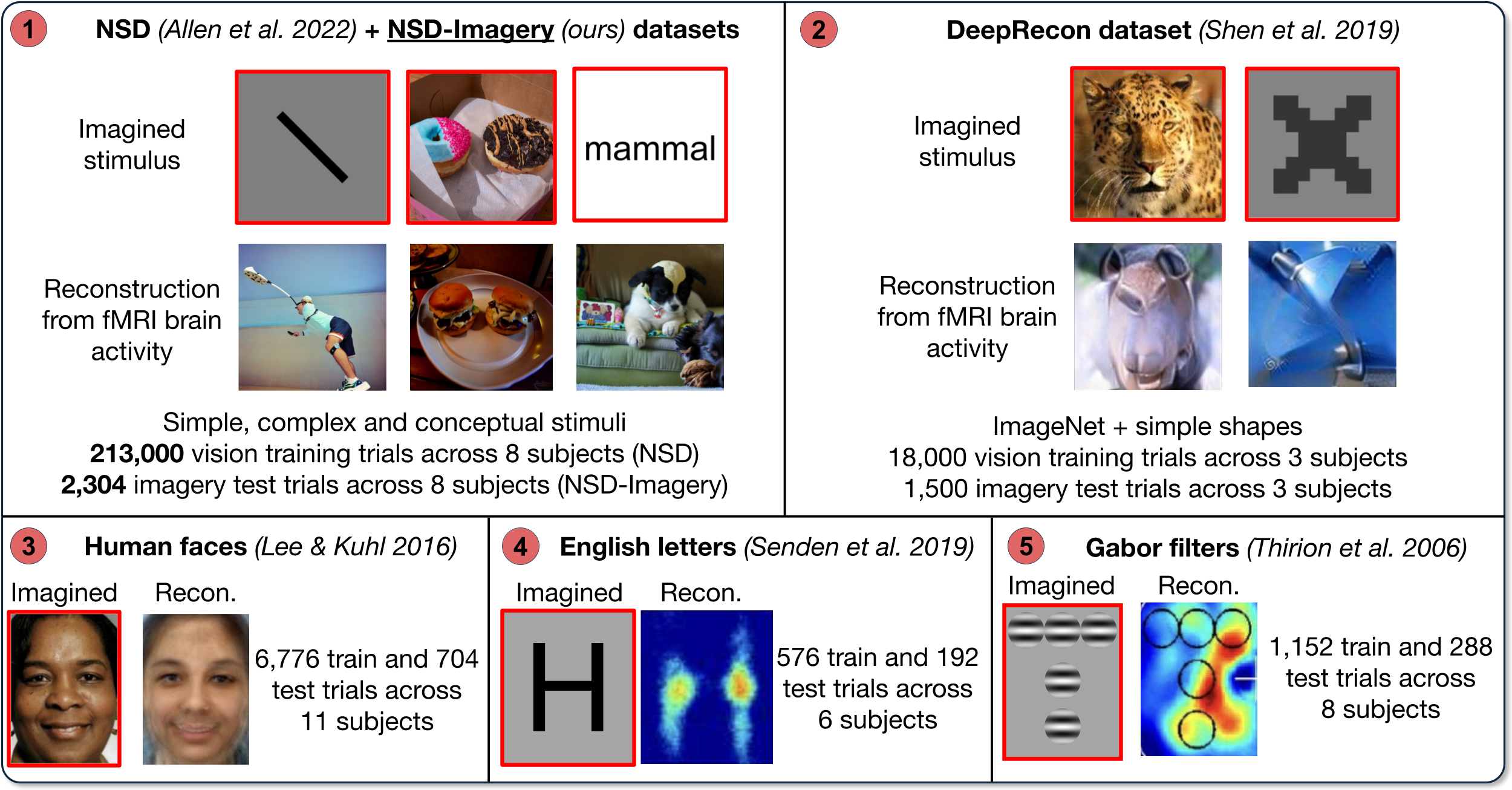}
\captionof{figure}{Overview of our dataset (NSD-Imagery) against 
previous imagery reconstruction datasets. $1$: NSD \cite{allen_massive_2022} and NSD-Imagery (ours), reconstructions of mental images are produced by Brain Diffuser \cite{ozcelik2023braindiffuser}. $2$: DeepRecon \cite{shen_deep_2019}, reconstructions of mental images produced by \citet{KOIDEMAJIMA2024349}. $3$: Human faces (\citet{lee_reconstructing_2016}). $4$: English letters (\citet{Senden_Emmerling_van}), reconstructions produced by \citet{goebel_reading_2022}. $5$: Gabor filters (\citet{thirion_inverse_2006}).
\label{fig:header}}
\end{strip}

\begin{abstract}
We release NSD-Imagery, a benchmark dataset of human fMRI activity paired with mental images, to complement the existing Natural Scenes Dataset (NSD), a large-scale dataset of fMRI activity paired with seen images that enabled unprecedented improvements in fMRI-to-image reconstruction efforts. Recent models trained on NSD have been evaluated only on seen image reconstruction. Using NSD-Imagery, it is possible to assess how well these models perform on mental image reconstruction. This is a challenging generalization requirement because mental images are encoded in human brain activity with relatively lower signal-to-noise and spatial resolution; however, generalization from seen to mental imagery is critical for real-world applications in medical domains and brain-computer interfaces, where the desired information is always internally generated. We provide benchmarks for a suite of recent NSD-trained open-source visual decoding models (MindEye1, MindEye2, Brain Diffuser, iCNN, Takagi et al.) on NSD-Imagery, and show that the performance of decoding methods on mental images is largely decoupled from performance on vision reconstruction. We further demonstrate that architectural choices significantly impact cross-decoding performance: models employing simple linear decoding architectures and multimodal feature decoding generalize better to mental imagery, while complex architectures tend to overfit visual training data. Our findings indicate that mental imagery datasets are critical for the development of practical applications, and establish NSD-Imagery as a useful resource for better aligning visual decoding methods with this goal.

\end{abstract}
\vspace{-15pt}
\section{Introduction}

The ability to decode mental states from brain activity is a longstanding goal of neuroscience. Mental images--visual representations not driven by retinal input--are an especially appealing target for decoding since an externalized mental image could, in principle, depict information stored in brain activity patterns that is difficult or impossible to read out by other means (e.g., by speaking). Access to this information could help researchers better understand cognitive processes; it could also be useful in a clinical setting \cite{pearson2015mental}, where millions of patients are left unable to communicate through conventional means as a result of traumatic brain injuries, and many common afflictions manifest as a profound dysregulation of unwanted or confusing mental imagery \cite{holmes2010mental}.

Most previous attempts to classify, retrieve, or reconstruct mental images have used a cross-decoding approach in which an encoding or decoding model is trained on brain activity evoked by seen images and then tested on brain activity recorded during mental imagery \cite{thirion_inverse_2006,shen_deep_2019,KOIDEMAJIMA2024349}. This approach is strongly justified by basic neuroscience that has demonstrated the extensive overlap in the representation of seen and mental images \cite{kosslyn2006case,stokes_top-down_2009,goebel_reading_2022,naselaris_voxel-wise_2015,reddy_reading_2010}. Excitement about the prospects offered by cross-decoding approaches has increased significantly of late, due to the rapid and dramatic improvement in vision decoding methods (e.g.,  \cite{Scotti2024MindEye2, scotti_reconstructing_2023}).

It has been unclear if recently developed methods for decoding seen images might generalize to mental imagery, which is encoded in brain activity with much lower signal-to-noise ratios than vision \cite{imagerysnr} and at a lower spatial resolution in the early visual areas that represent much of the structural detail of seen images \cite{BREEDLOVE20202211}. In particular, it is not known whether modern vision decoding methods can yield reconstructions that naive human observers would accurately identify as corresponding to the target images, a benchmark recently achieved by reconstructions of seen images \cite{Scotti2024MindEye2} and a minimum requirement for practical applications.

It has also been an open question whether the complexity of the imagined target stimuli might limit the quality of reconstructions of mental images. Although previous works (see Figure~\ref{fig:header}) have demonstrated reconstructions of simple imagined stimuli such as blobs \cite{thirion_inverse_2006}, letters \cite{goebel_reading_2022,Senden_Emmerling_van}, and single natural objects \cite{shen_deep_2019, KOIDEMAJIMA2024349,lee_reconstructing_2016}, there is no precedent for reconstructing complex natural scenes with multiple objects. To address these gaps, we make three main contributions:

\begin{enumerate}
    \item We release the benchmark dataset NSD-Imagery. This dataset is an extension to the Natural Scenes Dataset (NSD) \cite{allen_massive_2022}, which is a large-scale dataset of fMRI activity paired with seen images that is the current standard used to train brain-to-image reconstruction models \cite{scotti_reconstructing_2023, Scotti2024MindEye2}. NSD-Imagery provides held-out test trials where the same NSD participants performed a mental imagery task across varying stimulus complexity, thereby allowing researchers to test the generalization capabilities of vision decoding methods trained on the core NSD data. 
    \item We demonstrate that while the reconstruction performance of \textit{individual stimuli} is correlated between vision and mental imagery, \textit{methods} with improved performance on vision decoding do not necessarily produce improved performance for mental imagery decoding—at least for the handful of high-performing vision reconstruction methods tested here—a finding with critical importance to efforts developing methods that perform well on downstream tasks requiring mental images.
    \item We conduct extensive analyses using human raters and demonstrate that, despite the nuance in the previous point, some recent vision decoding methods do generalize to mental images, a promising finding that demonstrates the utility of using contemporary vision decoding models to improve reconstructions of mental images.
    
\end{enumerate}

\section{Related work}
The open releases of deep learning models, such as CLIP \cite{radford2021learning} and Stable Diffusion \cite{stablediffusion}, as well as large-scale fMRI datasets like NSD \cite{allen_massive_2022} where tens of thousands of images were shown to human subjects, sparked a surge of research papers showing high-quality reconstructions of seen images from human brain activity \cite{takagi2022_decoding,takagi2023improving,ozcelik_reconstruction_2022,ozcelik2023braindiffuser,ferrante_brain_2023,gaziv_self-supervised_2022,scotti_reconstructing_2023,kneeland2023reconstructing,kneeland_second_2023,kneeland_brain-optimized_2023,ferrante_through_2023,thual_aligning_2023,chen_cinematic_2023,chen_seeing_2023,sun_contrast_2023,mai_unibrain_2023,xia_dream_2023}. 
In contrast to earlier efforts \cite{naselaris_bayesian_2009, shinji_reconstruct, kamitani_decoding_2005,st-yves_generative_2018,shen_deep_2019,shen_end--end_2019,seeliger_generative_2018,lin_dcnn-gan_2019, gu_decoding_2023}, these methods map fMRI brain activity patterns to embeddings of pretrained deep learning models that are used to drive a diffusion model \cite{vdvae,podell_sdxl_2023,xu_versatile_2023,Scotti2024MindEye2} to generate image reconstructions of the content present in visual cortex. 

Generalizing these vision decoding methods to mental imagery is challenging because of the many documented differences between vision and mental imagery in the human brain\cite{lee2012disentangling, pearson2019human}. When compared to vision, brain activity patterns measured during mental imagery have much lower signal-to-noise ratios (SNR)  \cite{imagerysnr}, vary along fewer signal dimensions \cite{saharoycompressed}, and encode imagined stimuli with expanded receptive fields and lower spatial frequency preferences, especially in early visual cortex \cite{BREEDLOVE20202211}. These findings indicate that the spatial resolution of mental imagery is reduced \cite{favila2019spatial}, and that reconstructions of mental images are likely to be less structurally consistent than reconstructions of seen images \cite{dijkstra2024uncovering}. Previous works have also shown differences between vision and imagery in functional connectivity \cite{dijkstra2017distinct} and in temporal dynamics \cite{dijkstra2018differential}.

Despite these challenges, previous studies have reported success at decoding mental imagery from human brain activity. Many fMRI studies have reported accurate $n$-way classification of mental images \cite{cichy2012imagery, lee2012disentangling, albers2013shared}. Accurate retrieval of mental images of natural scenes using a visual encoding model was first reported in \citet{naselaris_voxel-wise_2015}. A preliminary attempt at retrieval for mental images in the NSD-Imagery dataset was reported in \citet{styvespredictmi}. To our knowledge, the first study to reconstruct mental images encoded in human brain activity measured with fMRI was \citet{thirion_inverse_2006}, in which the imagined target stimuli were simple configurations of blobs. Subsequent studies reported mental imagery reconstructions using simple stimuli such as blobs,
letters, or singular natural objects \cite{goebel_reading_2022,Senden_Emmerling_van,lee_reconstructing_2016,shen_deep_2019, KOIDEMAJIMA2024349}. 

Our work is distinguished from past work in the following ways: (1) we release a dataset that allows for cross-decoding mental images using NSD, the largest currently available vision dataset; (2) we study the generalization from perception to mental imagery of state-of-the-art (SOTA) vision reconstruction methods trained on NSD; (3) we directly compare the relative limitations of reconstructing stimuli of different distributions and levels of complexity; (4) we conduct large-scale tests of reconstruction quality with human raters; and (5) we quantify the correlation between the quality of seen image reconstructions and mental image reconstructions across stimuli and different reconstruction methods.

\section{NSD-Imagery Dataset}
\label{datasets}

The NSD-Imagery dataset extends NSD by incorporating a small set of 7T functional magnetic resonance imaging (fMRI) responses collected from the same subjects during mental imagery tasks. This enables the evaluation of vision decoding models on internally generated visual representations that are more relevant to future downstream use cases of brain decoding models. For this extension, all $8$ participants from the original NSD study underwent an additional scanning session, following the same high-resolution fMRI acquisition protocols as NSD.

\begin{figure}[!htb]
\vspace{-8pt}
\setlength{\abovecaptionskip}{2pt}    
\setlength{\belowcaptionskip}{-15pt}    
\includegraphics[width=\columnwidth]{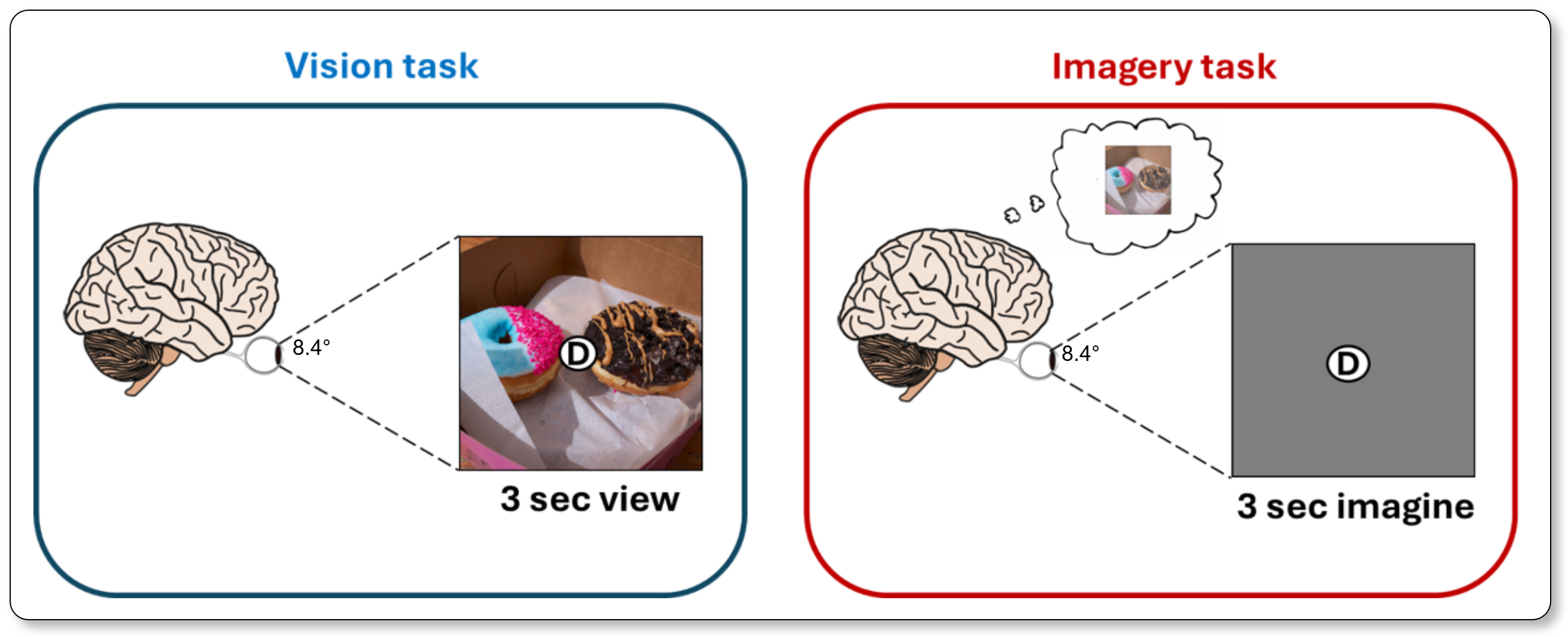}
\caption{Overview of the tasks utilized for the NSD-Imagery benchmark.} 
\label{figure:dataset}
\end{figure}

Each participant completed runs of three task types (vision, imagery, and attention) performed with three sets of target stimuli (simple, complex, and conceptual), resulting in 9 distinct run types, of which the imagery runs were completed twice for a total of 12 runs. Each run contained $8$ repetitions of the $6$ stimuli in each set, for a total of 576 trials per subject. The vision and imagery tasks (Figure \ref{figure:dataset})  are the primary tasks utilized for the NSD-Imagery benchmark. The target stimuli were carefully designed to encompass varying levels of complexity:
\begin{enumerate}[label=(\Alph*)]
    \item \textit{Simple stimuli}: Six simple geometric shapes--four oriented bars (0°, 45°, 90°, 135°) and two crosses ("+" and "×")--all constructed from black bars on a gray background.
    \item \textit{Complex stimuli}: Five natural scenes selected from the NSD shared1000 and one artwork (``The Two Sisters" by Kehinde Wiley). The natural scene images were chosen based on a recognizability score derived from participants' performance in the original NSD sessions, ensuring a range of familiarity and visual content.
    \item \textit{Conceptual stimuli}: Six single-word concepts describing abstract visual features or objects (e.g., stripes, banana, mammal), rather than specific images. The \textit{vision} trials for conceptual stimuli showed varying images of natural scenes corresponding to the target concept for each trial. Since the value of decoding a trial-averaged response to multiple visual stimuli is not clear, we recommend excluding the vision trials of conceptual stimuli from evaluations on the benchmark, and we present results here only on the imagery trials of conceptual stimuli.
\end{enumerate}

Each stimulus was associated with a unique single-letter cue, and participants memorized all 18 cue-stimulus pairs prior to scanning. Pre-scan practice sessions were conducted to ensure familiarity with the cues and stimuli, involving both visual presentations and verbal recall to reinforce memory.

In the \textit{vision runs}, participants were presented with images accompanied by the corresponding letter cue displayed at the central fixation point. Each trial lasted 4 seconds, consisting of 3 seconds of image and cue presentation, followed by a 1-second rest period with only the fixation cross displayed. Participants indicated via button press whether the presented image matched the letter cue. Images were displayed within a square frame occupying 8.4° × 8.4° of the subject's field of view, consistent with NSD stimuli presentations.

In the \textit{imagery runs}, participants were shown only the cue letter at the fixation point within the same frame used in vision runs, and were instructed to imagine the corresponding stimuli. Each trial also lasted 4 seconds, with 3 seconds allocated to vividly imagine the cued target stimulus, projecting the mental image onto the space outlined by the frame. This was followed by a 1-second rest period. Participants rated the vividness of their mental image via button press, pressing one of two buttons to indicate "vivid" or "not vivid."

In the \textit{attention runs}, participants were given a letter cue and asked to detect whether the cued stimulus was present in a series of rapidly presented images. While these runs are released as part of the NSD-Imagery dataset, their value for assessing vision reconstruction methods is unclear, and we recommend excluding these from evaluations on the NSD-Imagery benchmark.
More information on NSD-Imagery and details on data preprocessing are in Appendix \ref{app:datainfo}

\subsection{Dataset Release}
Instructions for accessing the NSD-Imagery dataset can be found at \href{www.naturalscenesdataset.org}{www.naturalscenesdataset.org}.

\section{Results}
\subsection{Evaluation Methods}
\label{decodingmethods}
In this study, we test a suite of $5$ fMRI-to-Image models, including Brain Diffuser \cite{ozcelik2023braindiffuser}, MindEye1 \cite{scotti_reconstructing_2023}, MindEye2 \cite{Scotti2024MindEye2}, iCNN \cite{shen_deep_2019}, and the +Decoded Text method proposed by Takagi et al. \cite{takagi2022_decoding, takagi2023improving}. We utilize the authors' open-source implementations for all vision decoding methods shown in this paper to generate reconstructions on NSD-Imagery. All reconstruction methods tested in this paper were trained exclusively on NSD \cite{allen_massive_2022}, the largest currently available neuroimaging dataset for vision. Only subjects $1$, $2$, $5$, and $7$ completed the full NSD experiment, so reconstruction methods generally train only on them.
Stimulus images were sampled from the Common Objects in Context (COCO) database \cite{microsoftcoco} (CC BY 4.0), which contains a diverse set of complex natural scenes.

\subsection{Reconstructions of seen images}
\label{visionbaseline}
The $5$ selected decoding methods were first applied to the $12$ stimuli available in the vision trials of the NSD-Imagery benchmark. $10$ reconstructions were sampled from the posterior distribution of each method. Best-case reconstructions from each method can be seen in Figure \ref{figure:vision}. Median and worst-case reconstructions can be seen in Appendix \ref{app:median_worst}.
Complete quantitative results from all methods on the vision trials can be seen in the middle of Table \ref{table:combined}. 

\begin{figure}[!htb]
\setlength{\abovecaptionskip}{3pt}    
\setlength{\belowcaptionskip}{-15pt}
\centering
\includegraphics[width=0.84\columnwidth]{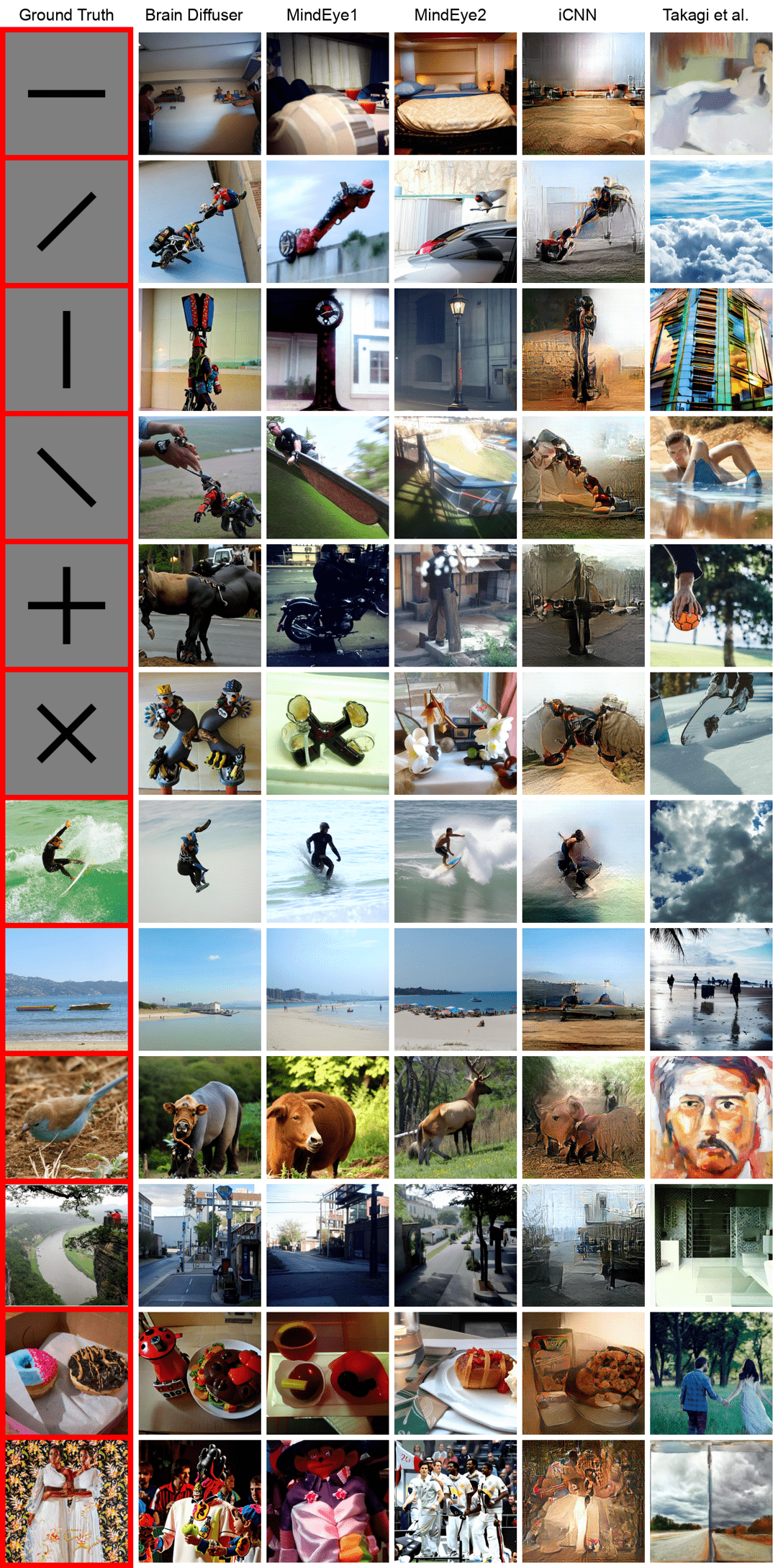}
\caption{Qualitative comparison of reconstruction methods on \textit{simple} and \textit{complex stimuli} seen during vision. 
Reconstructions selected for the figure are the best samples for each method and stimuli as assessed by quantitative performance across all metrics in Table \ref{table:combined}.} 
\label{figure:vision}
\end{figure}

In Figure \ref{figure:vision}, we observe reconstructions that appear qualitatively similar to the reconstructions reported in the original method publications. 
The NSD-Imagery dataset presents an interesting test of model generalization to new stimuli, as the simple stimuli are well outside the distribution of complex naturalistic images the models were trained on. 
The decoding models enforce this prior on the outputs, which produces simple stimuli reconstructions that are structurally coherent with the ground truth orientation, but with seemingly nonsensical semantic contents and fine details. 
While these out-of-distribution stimuli soften some recent criticisms of brain decoding models trained on NSD \cite{Shirakawa2024SpuriousRF}, they also illustrate the limitations of the training prior on the reconstruction process, as it can be difficult to generalize to out-of-distribution images and stimuli types. We note that out-of-distribution generalization is a topic that can be explored in greater detail using NSD-Synthetic \cite{nsdsynthetic}, a recently released NSD extension that includes responses to carefully controlled synthetic images. 

\subsection{Reconstructions of mental images}
\label{extendingimagery}
The $5$ decoding methods were next applied to the $18$ stimuli available in the mental imagery trials of the NSD-Imagery benchmark. $10$ reconstructions were sampled from the posterior distribution of each method.
Best-case reconstructions from each method can be seen in Figure \ref{figure:imagery}. Median and worst-case reconstructions can be seen in Appendix \ref{app:median_worst}. Complete quantitative results from all methods on the mental imagery trials can be seen at the top of Table \ref{table:combined}.

\begin{figure}
\setlength{\abovecaptionskip}{3pt}
\centering
\includegraphics[height=0.95\textheight]{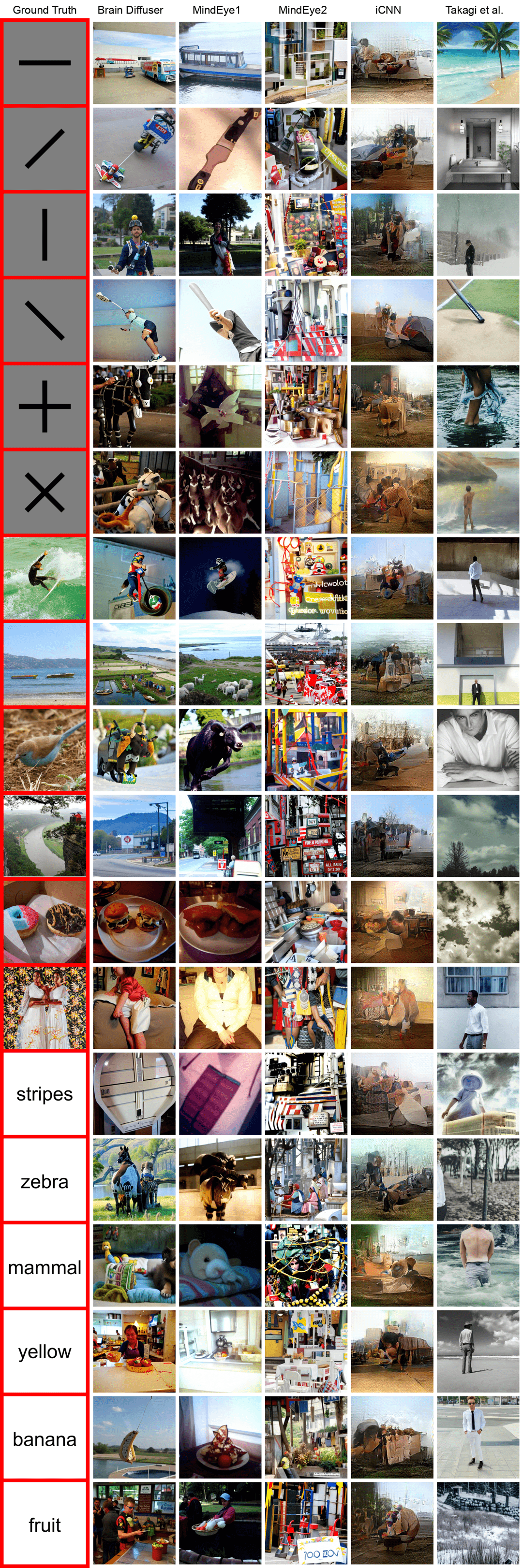}
\caption{Qualitative comparison of reconstruction methods on all $18$ imagined stimuli from NSD-Imagery. Samples are selected the same way as Figure \ref{figure:vision}.} 
\label{figure:imagery}
\end{figure}

At a glance, several methods appear to generalize well to mental imagery, successfully reconstructing images that qualitatively reflect similar categories and themes for the \textit{complex stimuli}, such as food, water sports, animals, and outdoor scenes.  For \textit{simple stimuli}, the overall structure and orientation of the ground truth images are largely preserved in the mental image reconstructions. The \textit{conceptual stimuli} also produce interesting reconstructions, in many cases demonstrating noticeable correspondence to the target concept, although it is difficult to quantify accuracy for these stimuli, as there are no ground truth images, and the precise content of the subject's mental images is not known.
Overall the reconstructions demonstrate substantial fidelity with respect to the content present in the ground truth images, and mark a major leap forward in mental image reconstruction compared to the current baseline introduced in \citet{KOIDEMAJIMA2024349} (Figure \ref{fig:header}).

\subsection{Feature metric evaluations of reconstruction quality}
\label{featuremetrics}
Several reconstruction methods demonstrate promisingly robust quantitative outcomes when applied to mental images, with MindEye1 and Brain Diffuser standing out as the top performers. All the methods do show a drop in quantitative benchmarks when reconstructing mental images (Table \ref{table:combined}, top) compared to the NSD-Imagery vision trials (Table \ref{table:combined}, middle) and the test set from NSD (Table \ref{table:combined}, bottom), an expected result given previous research demonstrating that brain activity responses to mental images have lower SNR and fewer dimensions of signal variance than vision \cite{imagerysnr, saharoycompressed}.
Interestingly, brain correlation scores in early visual cortex experience a much steeper drop than brain correlation scores in higher visual cortex, aligning with findings of broader receptive fields during imagery for voxels within early visual regions \cite{BREEDLOVE20202211}. 
This drop in spatial resolution is also reflected in the low-level metrics--which generally privilege accurate structure--as they experience a larger drop than higher-level metrics that are more tuned to semantic qualities such as object class. All of the methods evaluated also observe a slight drop in quantitative metrics on the vision trials of NSD-Imagery (Table \ref{table:combined}, bottom) compared to the test set from NSD, likely a result of the shift in image distribution for the simple stimuli.

\begin{table*}
    \setlength{\abovecaptionskip}{2pt}    
    \setlength{\belowcaptionskip}{-14pt}    
    \centering
    \setlength{\tabcolsep}{4pt}
    \small
    \resizebox{\textwidth}{!}{
    \begin{tabular}{lccccccccccc}
        \toprule
        Method & \multicolumn{2}{c}{Low-Level} & \multicolumn{4}{c}{High-Level} & \multicolumn{2}{c}{Distance} & \multicolumn{3}{c}{Brain Correlation} \\
        \cmidrule(lr){2-3} \cmidrule(lr){4-7} \cmidrule(lr){8-9} \cmidrule(l){10-12}
        & PixCorr $\uparrow$ & SSIM $\uparrow$ & Alex(2) $\uparrow$ & Alex(5) $\uparrow$ & Incep $\uparrow$ & CLIP $\uparrow$ & Eff $\downarrow$ & SwAV $\downarrow$ & Early Vis. $\uparrow$ & Higher Vis. $\uparrow$ & Visual Cortex $\uparrow$ \\
        \midrule
        \multicolumn{12}{c}{\textbf{NSD-Imagery Mental Imagery Trials}} \\
        \midrule
        \textbf{MindEye1 \cite{scotti_reconstructing_2023}} & \underline{0.086} & 0.349 & \textbf{59.56\%} & \textbf{61.00\%} & \underline{52.03\%} & \textbf{54.72\%} & \underline{0.948} & \underline{0.564} & \textbf{0.180} & \textbf{0.135} & \textbf{0.155} \\
        \textbf{Brain Diffuser \cite{ozcelik2023braindiffuser}} & 0.064 & 0.401 & \underline{52.14\%} & \underline{58.35\%} & \textbf{52.73\%} & \underline{54.07\%} & \textbf{0.935} & 0.585 & \underline{0.133} & \underline{0.127} & \underline{0.141} \\
        \textbf{iCNN \cite{shen_deep_2019}} & \textbf{0.108} & 0.340 & 50.57\% & 55.25\% & 49.39\% & 41.72\% & 0.994 & \textbf{0.560} & 0.113 & 0.062 & 0.113 \\
        \textbf{MindEye2 \cite{Scotti2024MindEye2}} & 0.036 & \underline{0.414} & 47.60\% & 55.38\% & 46.02\% & 50.78\% & 0.966 & 0.591 & 0.069 & 0.055 & 0.061 \\
        \textbf{Takagi et al. \cite{takagi2022_decoding,takagi2023improving}} & -0.006 & \textbf{0.455} & 41.88\% & 40.19\% & 43.26\% & 40.08\% & 0.976 & 0.606 & 0.000 & 0.004 & 0.000 \\
        \midrule
        \multicolumn{12}{c}{\textbf{NSD-Imagery Vision Trials}} \\
        \midrule
        \textbf{MindEye1 \cite{scotti_reconstructing_2023}} & \underline{0.218} & 0.412 & \textbf{73.56\%} & \underline{80.81\%} & \underline{62.44\%} & \underline{65.34\%} & \textbf{0.881} & \textbf{0.510} & \underline{0.374} & \textbf{0.253} & \underline{0.311} \\
        \textbf{Brain Diffuser \cite{ozcelik2023braindiffuser}} & 0.107 & \underline{0.455} & 60.34\% & 72.84\% & 60.95\% & 58.31\% & 0.908 & 0.555 & 0.247 & 0.229 & 0.255 \\
        \textbf{iCNN \cite{shen_deep_2019}} & \textbf{0.224} & 0.385 & \underline{71.67\%} & \textbf{81.35\%} & 61.16\% & 49.03\% & 0.926 & 0.524 & \textbf{0.442} & \underline{0.246} & \textbf{0.338} \\
        \textbf{MindEye2 \cite{Scotti2024MindEye2}} & 0.161 & \textbf{0.480} & 70.10\% & 77.52\% & \textbf{62.69\%} & \textbf{65.93\%} & \underline{0.886} & \underline{0.512} & 0.356 & 0.234 & 0.290 \\
        \textbf{Takagi et al. \cite{takagi2022_decoding,takagi2023improving}} & -0.013 & 0.412 & 41.55\% & 39.26\% & 39.26\% & 43.01\% & 0.969 & 0.610 & -0.006 & 0.016 & 0.009 \\
        \midrule
        \multicolumn{12}{c}{\textbf{NSD Shared1000 Test Set}} \\
        \midrule
        \textbf{MindEye1 \cite{scotti_reconstructing_2023}} & 0.319 & 0.360 & - & - & - & - & \underline{0.648} & \textbf{0.377} & 0.350 & \underline{0.377} & 0.378 \\
        \textbf{Brain Diffuser \cite{ozcelik2023braindiffuser}} & 0.273 & 0.365 & - & - & - & - & 0.728 & 0.421 & 0.353 & \underline{0.375} & \underline{0.381} \\
        \textbf{iCNN \cite{shen_deep_2019}} & \underline{0.321} & 0.336 & - & - & - & - & 0.797 & 0.528 & \textbf{0.410} & 0.371 & \textbf{0.395} \\
        \textbf{MindEye2 \cite{Scotti2024MindEye2}} & \textbf{0.322} & \textbf{0.431} & - & - & - & - & \textbf{0.619} & \textbf{0.344} & \underline{0.360} & 0.368 & 0.373 \\
        \textbf{Takagi et al. \cite{takagi2022_decoding,takagi2023improving}} & 0.246 & \underline{0.410} & - & - & - & - & 0.811 & 0.504 & 0.167 & 0.288 & 0.247 \\
        \bottomrule
    \end{tabular}
    }
    \caption{Quantitative comparison between reconstruction methods for NSD-Imagery using the simple and complex stimuli (Conceptual stimuli have no ground truth images). NSD Shared1000 test set results are reported at the bottom for comparison, with the exception of the 2-way comparison (2WC) metrics as these are not directly comparable to metrics on NSD-Imagery. PixCorr is the pixel-level correlation score. SSIM is the structural similarity index metric \cite{wang_image_2004}. AlexNet($2$) and AlexNet($5$) are the 2WCs of layers 2 and 5 of AlexNet \cite{alexnet}. CLIP is the 2WC of the output layer of the ViT-L/14 CLIP-Vision model \cite{radford2021learning}. Inception is the 2WC of the last pooling layer of InceptionV3 \cite{inceptionv3}. EffNet-B and SwAV are distance metrics gathered from EfficientNet-B13 \cite{tan_efficientnet_2020} and SwAV-ResNet50 \cite{caron_unsupervised_2021} models. For EffNet-B and SwAV distances, lower is better, while higher is better for all other metrics. Bold indicates best performance, and underline second-best performance. Additional notes on the metrics used, including explanations of 2-way comparisons and brain correlation scores, are in Appendix \ref{app:metrics}. A breakdown of model performance across the different types of stimuli is in Appendix \ref{app:stimtypes}.}
    \label{table:combined}
\end{table*}

\subsection{Human ratings of reconstruction quality}
\label{humanratings}
To be useful for applications, users, scientists, and clinicians will need to draw meaningful interpretations from the outputs of mental imagery decoding methods. Thus, human judgments of the quality of mental image reconstructions are important metrics of performance. Although the image feature metrics discussed in Section \ref{featuremetrics} may be a reasonable proxy for human judgment, numerous research efforts have established that these metrics do not always closely align with human subjective assessments of content \cite{peceptualsimilarity} or quality \cite{pickapic,Scotti2024MindEye2}, and a small dataset like NSD-Imagery can be particularly susceptible to such volatility. 
In light of this, we conducted several online behavioral experiments in which human raters (n=$500$) assessed the quality of the reconstructions. 
For detailed experiment protocols see Appendix \ref{app:behavioral}.

An initial image-identification experiment asked human raters to perform a two-alternative forced-choice judgment about whether a reconstruction was more similar to the ground truth image than a randomly selected reconstruction of a different stimulus within the same stimulus type.
Results from this experiment (Table \ref{table:2afc}) underscore promising generalizability for several vision decoding methods, notably MindEye1 and Brain Diffuser which produce strong accuracy rates for mental image reconstructions. We also note that MindEye2, the current SOTA vision decoding method, generalizes extremely poorly, producing near-chance performance. We hypothesize that this decoupling of vision and imagery decoding performance is a result of architectural design, discussed further in Section \ref{discussion}.

For us, the method from Takagi et al. underperforms on both the vision and imagery trials of NSD-Imagery, producing chance-level reconstructions as evaluated by both the image feature metrics in Table \ref{table:combined} and human raters in Table \ref{table:2afc}. For this reason, we have excluded it from further analyses.

\begin{table}[!htb]
\vspace{-15pt}
    \setlength{\abovecaptionskip}{3pt}    
    \setlength{\belowcaptionskip}{-20pt}    
    \centering
    \captionsetup{font=small}
    \setlength{\tabcolsep}{3pt} 
    \small
    \resizebox{\columnwidth}{!}{%
    \begin{tabular}{lcccc}
    \\
    \multicolumn{5}{c}{\textbf{Human Identification Accuracy}} \\
    \midrule
    Method & All Stimuli ↑ & Simple ↑ & Complex ↑ & Conceptual ↑ \\
    \midrule
    \multicolumn{5}{c}{\textbf{NSD-Imagery Mental Imagery Trials}} \\
    \midrule
    \textbf{MindEye1 \cite{scotti_reconstructing_2023}} & \underline{73.00\%} & \textbf{71.01\%} & \underline{82.28\%} & \underline{65.68\%} \\
    \textbf{Brain Diffuser \cite{ozcelik2023braindiffuser}} & \textbf{73.95\%} & \underline{68.20\%} & \textbf{82.70\%} & \textbf{71.01\%} \\
    \textbf{iCNN \cite{shen_deep_2019}} & 66.15\% & 66.81\% & 70.04\% & 61.60\% \\
    \textbf{MindEye2 \cite{Scotti2024MindEye2}} & 56.96\% & 50.21\% & 64.83\% & 55.74\% \\
    \textbf{Takagi et al. \cite{takagi2022_decoding,takagi2023improving}} & 52.89\% & 57.26\%  & 48.31\%  & 53.14\%  \\
    \midrule
    \multicolumn{5}{c}{\textbf{NSD-Imagery Vision Trials}} \\
    \midrule
    \textbf{MindEye1 \cite{scotti_reconstructing_2023}} & \textbf{84.29\%} & \textbf{79.32\%} & \underline{89.32\%} & \textit{N/A} \\
    \textbf{Brain Diffuser \cite{ozcelik2023braindiffuser}} & 80.13\% & 71.79\% & 88.46\% & \textit{N/A} \\
    \textbf{iCNN \cite{shen_deep_2019}} & 74.52\% & \underline{75.85\%} & 73.19\% & \textit{N/A} \\
    \textbf{MindEye2 \cite{Scotti2024MindEye2}} & \underline{83.05\%} & 75.54\% & \textbf{90.56\%} & \textit{N/A} \\
    \textbf{Takagi et al. \cite{takagi2022_decoding,takagi2023improving}} & 49.58\%  & 48.95\%  & 50.20\% & \textit{N/A} \\
    
    \bottomrule
    \end{tabular}
    }
    \caption{Human identification accuracy scores for the mental imagery and vision trials of the NSD-Imagery benchmark. Scores are provided for each method and stimulus type. The best value is bolded, and the second best is underlined. The chance threshold is 50\%. P-values for Takagi et al. \cite{takagi2022_decoding,takagi2023improving} are $0.891$ for vision and $0.133$ for imagery. P-values for all other methods are \textless{}$0.001$.}
    \label{table:2afc}
\end{table}

\subsection{Comparison of reconstruction quality for different stimulus types}
\begin{figure}[!htb]
\vspace{-5pt}
    \setlength{\abovecaptionskip}{-1pt}    
    \setlength{\belowcaptionskip}{-15pt}    
    \centering
        \includegraphics[width=\columnwidth]{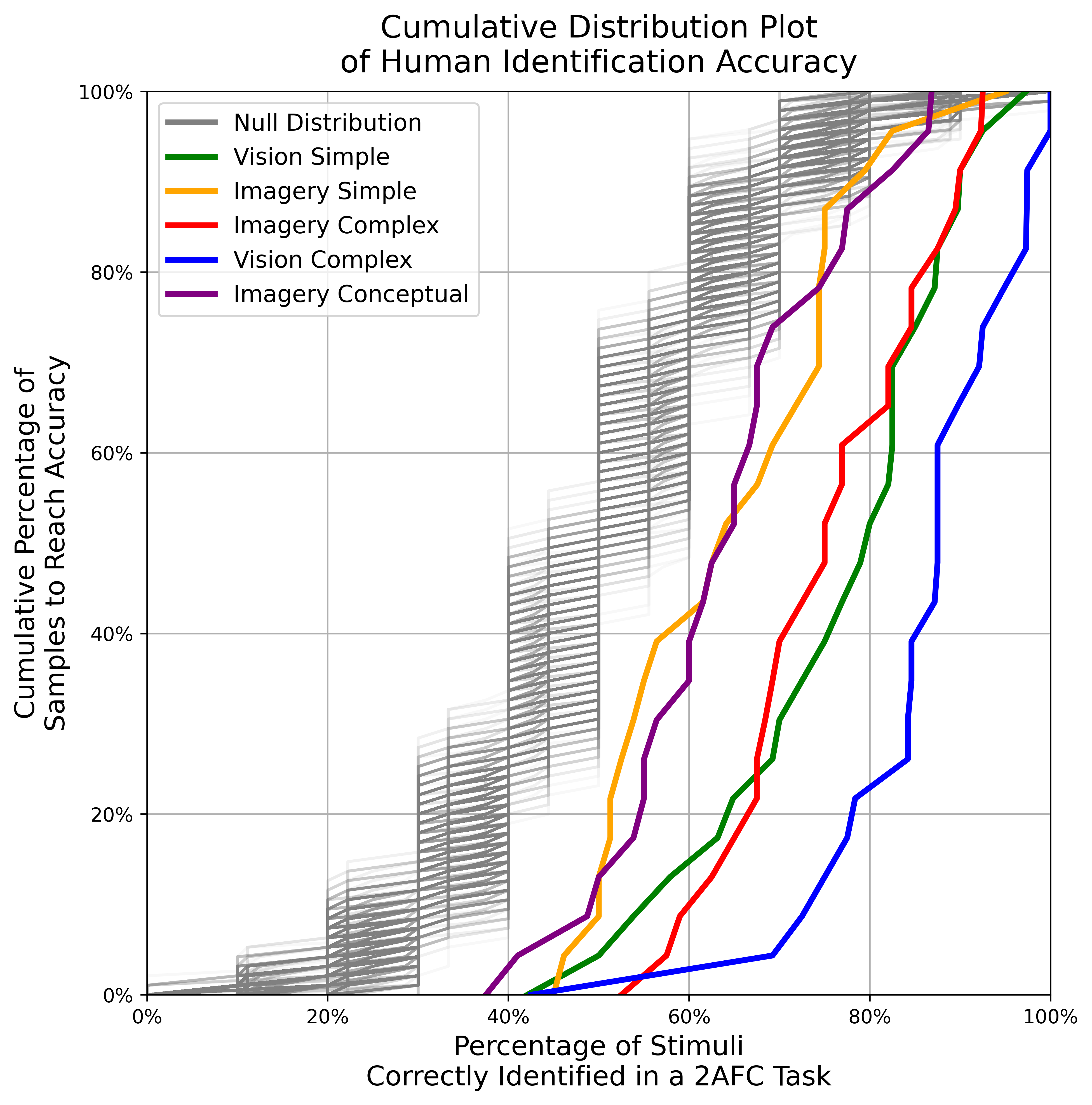}
    \caption{Cumulative distribution plot of human two-way identification accuracy across stimulus/modality types (line colors), collapsed across methods (excluding Takagi et al.). Data is from Table \ref{table:2afc}. The X-axis indicates intervals of human identification accuracy, and the Y-axis indicates the cumulative percentage of samples to meet the accuracy threshold at each interval. Less area under the curve is better. Null distribution is plotted in gray.} 
    \label{fig:exp1}
\end{figure}

To understand how reconstruction quality varies across inference modes, we calculated the cumulative distribution of human ratings from our first experiment (Figure \ref{fig:exp1}) by pooling across methods and conditioning on modality (vision/imagery) and stimulus complexity (simple/complex/conceptual). 
As anticipated, complex visual stimuli (blue line) yielded the strongest performance, reflecting the primary training scenario of these methods. Interestingly, under this rating method, reconstructions of \textit{imagined} complex stimuli (natural scenes) have very similar assessments to, and in some cases outperform, reconstructions of \textit{seen} simple stimuli (bars and crosses). 
We infer that the drop in performance for simple and conceptual stimuli is caused by their deviation from the distribution of the stimuli used for training, as well as the priors of the pre-trained diffusion model, which also carries a bias towards naturalistic stimuli. This suggests that decoding performance may be influenced by the consistency of training and test distributions and the priors of the pre-trained models more than by the decoded stimuli's inherent complexity (or other decoding challenges). 

\begin{figure}[!htb]
\setlength{\abovecaptionskip}{-1pt}    
\setlength{\belowcaptionskip}{-18pt}    
    \centering
    \includegraphics[width=\columnwidth]{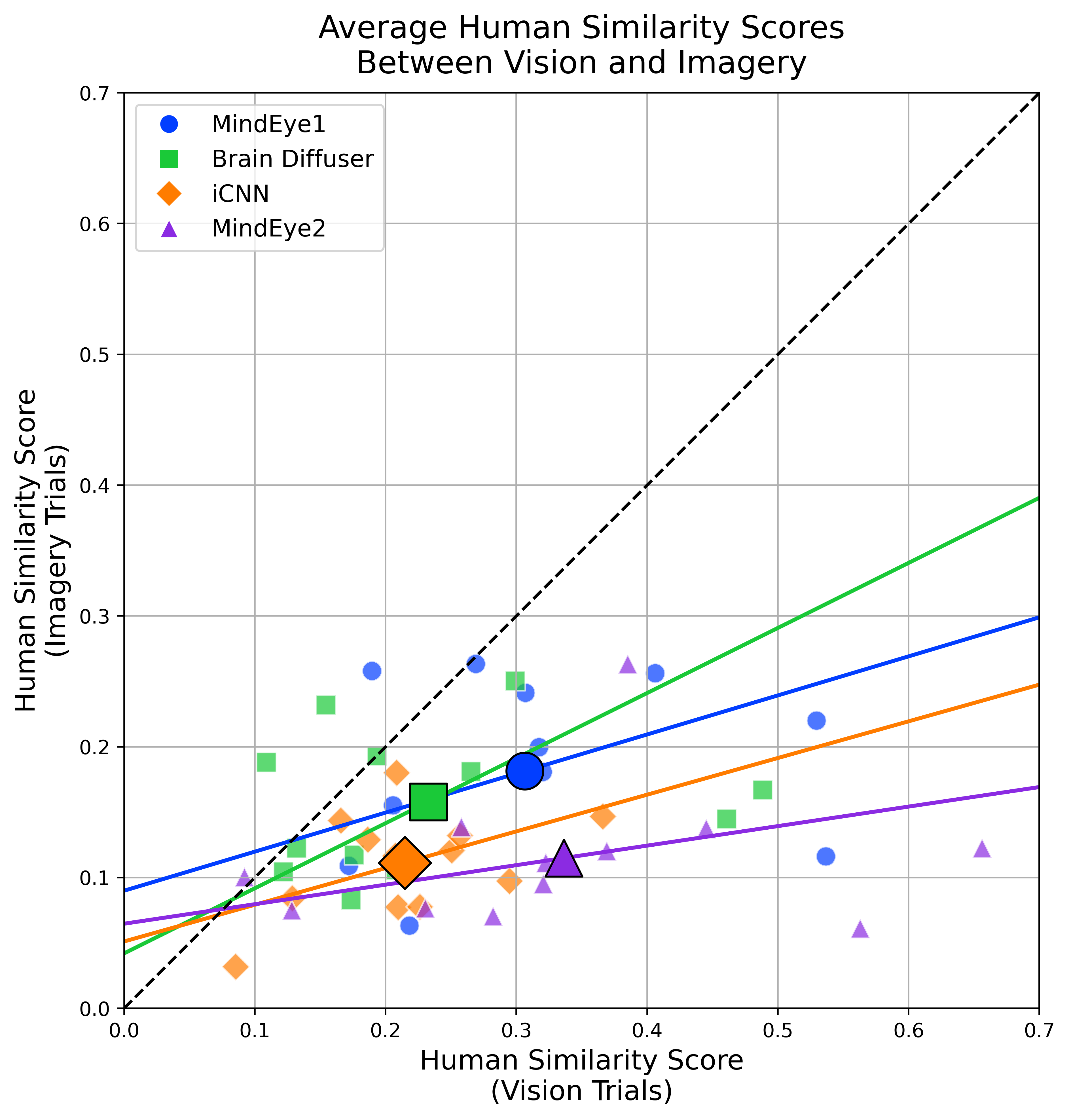}
    \caption{Human similarity score results for the simple and complex stimuli. The X-axis represents the similarity scores for the vision reconstruction, while the Y-axis represents the similarity scores for the imagery reconstruction. 
    The average response scores for the $12$ samples are plotted individually for each method, denoted with colors and shapes. 
    The average score of all $12$ samples is plotted with a larger bold point of the same shape and color. 
    A larger slope for the color-coded fitted lines represents better generalization performance between the imagery and vision trials. 
    The dashed line at unity represents equal performance between vision and imagery reconstructions.} 
\label{fig:exp3}
\end{figure}

\subsection{Correlation between seen and mental image reconstruction quality}
\label{similarityexperiment}

To obtain a direct comparison of the quality of matched seen and mental image reconstructions, we performed a second similarity-score experiment that presented human raters with a ground truth image, a reconstruction of that image from a vision trial, and a reconstruction of that image from an imagery trial, asking human raters to rate on a continuous scale how similar each reconstruction was to the ground truth image. We note that continuous similarity ratings like these can be extremely sensitive to the context of the samples being presented, and so while this experiment is useful for comparing the relative similarity between vision and imagery reconstructions, it is less informative of differences between decoding methods or other stimulus conditions.

We observe that these similarity scores of matched seen and mental image reconstructions (Figure \ref{fig:exp3}) were correlated across stimuli (MindEye1: r = $0.16$, Brain Diffuser: r = $0.19$, iCNN: r = $0.22$, MindEye2:  r = $0.13$; p $\textless{} 0.001$ for all methods, n = $480$ for each method), but we note that not all methods were equally correlated. In particular, MindEye2 again produces significantly weaker correlations than the older MindEye1, Brain Diffuser, and iCNN models.
The correlation between the similarity scores of vision and imagery reconstructions demonstrates that \textit{specific stimuli} that reconstruct well for vision tend to also reconstruct well for imagery, allowing us to infer that improvements to decoders of seen images will likely translate to improvements in decoding matched mental images, although the \textit{degree} to which these improvements translate appears to be heavily dependent on the specific architecture of the decoding model, and varies significantly between methods.

\section{Discussion}
\label{discussion}
We demonstrate the utility of the NSD-Imagery dataset in showing that, despite the massive loss of SNR during imagery relative to vision, several vision decoding methods based on recent advances in generative modeling robustly generalize to mental imagery without the need for domain-specific model training, and produce SOTA reconstructions of mental images. Previous SOTA reconstructions used the DeepRecon dataset, which faced fundamental limitations due to training data size (see Figure \ref{fig:header}). We hope our dataset will be useful to researchers training fMRI-to-Image models on NSD, enabling them to evaluate the generalizability of the models to out-of-distribution stimuli and mental images. 

The fact that variation in the vision reconstruction performance across the small set of high-performing models tested here does not explain the variation in mental image reconstruction performance across this same set of models demonstrates that the best vision decoding models are not necessarily the best mental imagery decoding models. Specifically, we observe that the high-dimensional ViT-bigG image embeddings used to drive the SDXL unCLIP generator in MindEye2 tend to break down when decoded from mental images, which have brain activity responses of much lower SNR. We also observe that the Brain Diffuser method appears to be the best of all the methods tested in generalizing to mental images, and hypothesize that this is likely due to its simple and robust ridge regression backbone, as well as the multi-modal image and text features employed in the architecture. Further analysis to find the optimal decoding architectures for generalizing vision decoding models to mental images would be rich ground for future work.

Our results show that in some cases, decoding performance on mental imagery for complex stimuli outperforms vision decoding performance for simpler stimuli that are out of distribution from the training images, indicating that target image complexity is not currently the primary limiting factor to mental image reconstruction quality. We infer from this result that alignment with the training distribution of the decoding model is currently a more relevant ceiling for the reconstruction of mental images than the raw complexity of the stimuli. Therefore, it seems likely that if vision reconstruction methods were trained on a broader distribution of stimuli, their performance for these stimuli would be greatly improved.

Our results indicate that training a vision decoder can serve as an effective proxy for directly optimizing a mental image decoder (although it is critical to evaluate vision decoders against the kind of data that NSD-Imagery provides). This is good news, as brain activity in response to seen images is far easier to measure and scale than activity related to mental imagery. Furthermore, reconstructions of seen images are a more tractable inference problem than reconstructions of mental images, as the "true" mental image may differ from the cued stimulus in ways that are difficult to predict, given our current understanding of mental imagery.
 
\subsection{Broader impacts}
\label{broaderimpacts}

We envision many applications for mental image reconstructions, including the development of diagnostic instruments for psychiatric conditions where imagery is intrusive or associated with strong negative emotions \cite{holmes2010mental}, and alternative communication methods for patients unable to communicate through conventional means. 

For the $50$ million unresponsive patients who suffer traumatic brain injuries every year \cite{50mtbipatients}, recovery of consciousness is a key milestone that predicts long-term functional recovery \cite{Giacino_Kalmar_1997}.
It is estimated that $15-20\%$ of them may have levels of consciousness not apparent through standard behavioral assessments, remaining undetected or covertly conscious \cite{consciousdetection}. 
Leveraging non-invasive brain imaging technologies to decode verifiable reconstructions of a patient's imagination could help bridge this gap, aiding in accurate diagnoses that could prevent the premature discontinuation of life support, a fate met by $70\%$ of brain injury patients who die in the hospital \cite{tbimortality}.

\subsection{Limitations and future work}
\label{limitations}
The primary limitation of research reconstructing mental images is the current size of mental imagery datasets. 
The NSD-Imagery dataset analyzed in this work contains only a small set of $18$ stimuli and thus does not allow for large-scale model training on the mental imagery data, necessitating a cross-decoding approach. 
The collection of large-scale mental imagery datasets remains challenging due to the complex experimental protocols required, as collecting imagery trials requires subjects to memorize stimulus images in advance of the scanning session, limiting the number of stimuli that can be presented. As such, the collection and development of comprehensive mental imagery datasets is a critical area for future development and investment. 

\subsection{Ethical considerations}
\label{ethics}
There is a rapidly increasing number of research efforts aiming to decode thoughts, mental images, and other internal representations from brain activity. While there are clear downstream benefits to this research, it also raises significant questions regarding its broader societal impacts and potential for misuse. As this technology develops, we believe it is important to begin developing a legal and ethical framework for the application of brain-decoding devices that rigorously safeguards users and ensures that the technology is deployed transparently, responsibly, and for the benefit of humankind.


%% file: appendix.tex
\appendix
\section{Appendix}

\subsection{Additional dataset information}
\label{app:datainfo}
Stimuli within both NSD \cite{allen_massive_2022} and NSD-Imagery were displayed at $8.4\times 8.4$ degrees. All fMRI data in the NSD were collected at ultra-high field (7T) using a whole-brain, $1.8$-mm, $1.6$-s, gradient-echo, echo-planar imaging (EPI) pulse sequence. The fMRI responses are expressed in terms of ``betas'' ($\beta$; each $\beta$ is a measure of the amplitude of BOLD signal evoked by a single image in a single voxel) obtained from a general linear model (GLM) analysis. Betas indicate BOLD response amplitudes evoked by each stimulus trial relative to the baseline signal level present during the absence of a stimulus (gray screen). All reconstruction methods evaluated in this work were trained using GLMsingle results provided with the NSD data release, and evaluated on the GLMsingle preparations of the NSD-Imagery data, specifically, the $1.8$-mm volume preparation of the data and version $3$ of the GLM betas (betas\_fithrf\_GLMdenoise\_RR).

The shared1000 test set—which reconstruction methods are typically evaluated on—was sampled from scanning sessions of NSD where training data was also collected. Thus, cross-session non-stationarities are likely to have had similar impacts on the training and evaluation data. The vision trials in NSD-Imagery, however, were collected during a separate scanning session. Thus, it is likely that cross-session non-stationarities had a more detrimental impact on decoder performance when generalizing to NSD-imagery than to the shared1000 data. This could explain why the performance of all decoders on the NSD-Imagery vision trials were generally lower than previously published results for the shared1000 trials. 
The drop in performance relative to the shared1000 could also have a number of other causes, including a different task being used in the NSD-Imagery vision trials (cue-matching vs continuous recognition), a different number of stimulus repetitions in NSD-Imagery ($8$ repetitions vs. $3$ repetitions), or the possibility that the GLMsingle algorithm used to preprocess the fMRI responses \cite{prince_improving_2022} is less effective on small datasets. 

Across all methods evaluated in this paper \cite{scotti_reconstructing_2023, Scotti2024MindEye2, ozcelik2023braindiffuser, kneeland_brain-optimized_2023, takagi2022_decoding, takagi2023improving}, 
we trained the models using the full 40 sessions of the NSD training data (not including the shared1000), and we perform inference using the same brain region as the original paper authors. For the majority of methods evaluated in the paper, this means we utilize only voxels from the “nsdgeneral” brain region, defined by the NSD authors as the subset of voxels in posterior cortex most responsive to the visual stimuli presented (between 13,000 to 16,000 voxels per participant). The method from Takagi et al. is the only method to deviate from this, instead using the respective ROIs for early and higher (ventral) visual regions included in the streams atlas of the NSD. 

Because the NSD-Imagery dataset comprises both vision and imagery trials within a single scanning session, as well as multiple types of discrete stimulus types, we Z-scored the fMRI data within each experimental run separately. A run is defined by a series of consecutive trials comprising the same visual modality and stimulus type (e.g., vision trials comprising simple stimuli), typically lasting 4 minutes.  Normalizing trials for vision and imagery runs separately provides some control against non-stationary brain activity (e.g., changes in SNR) across imagery and vision.

\subsection{Additional evaluation metric details}
\label{app:metrics}
All metrics calculated in Table $1$ of the manuscript were calculated across 10 reconstructions sampled from the posterior distribution of each decoding method. A two-way comparison evaluates whether the feature embedding of the stimulus image is more similar to the feature embedding of the target reconstruction, or the feature embedding of a randomly selected "distractor" reconstruction. Two-way identification refers to percent correct across a set of two-way comparisons performed on a pool of distractor images. The two-way identification metrics we report, which are calculated using reconstructions of the $11$ other NSD-Imagery stimuli as distractors, are notably different from the two-way identification metrics presented in individual reconstruction papers that perform evaluations using reconstructions of the shared1000 as the pool of distractors. The pool of distractor images for NSD-Imagery is much smaller, and contains multiple distinct types of stimuli that may significantly alter the resulting identification accuracy metrics. Because of this difference, the two-way identification accuracy numbers are not directly comparable to two-way identification results evaluated on the shared1000 in other papers, and we do not report the shared1000 2WC metrics in Table $1$ of the manuscript. Brain correlation scores are the Pearson correlation between the averaged measured brain response $\beta$ and the predicted brain response $\beta'$ produced by a brain encoding model (GNet \cite{St-Yves_heirarchy}) averaged across voxels within a respective ROI in visual cortex, including the whole visual cortex, early visual cortical regions V1, V2, V3, and V4, and higher visual areas (set complement of visual cortex and early visual cortex). All metrics in Tables $1$, $2$, and $3$ were calculated and averaged across 10 images sampled from the output distribution of each method using a random seed. 

\clearpage
\subsection{Median and worst case reconstructions}
\label{app:median_worst}
\FloatBarrier
\begin{figure}[!htb]
\centering
\includegraphics[width=0.87\columnwidth]{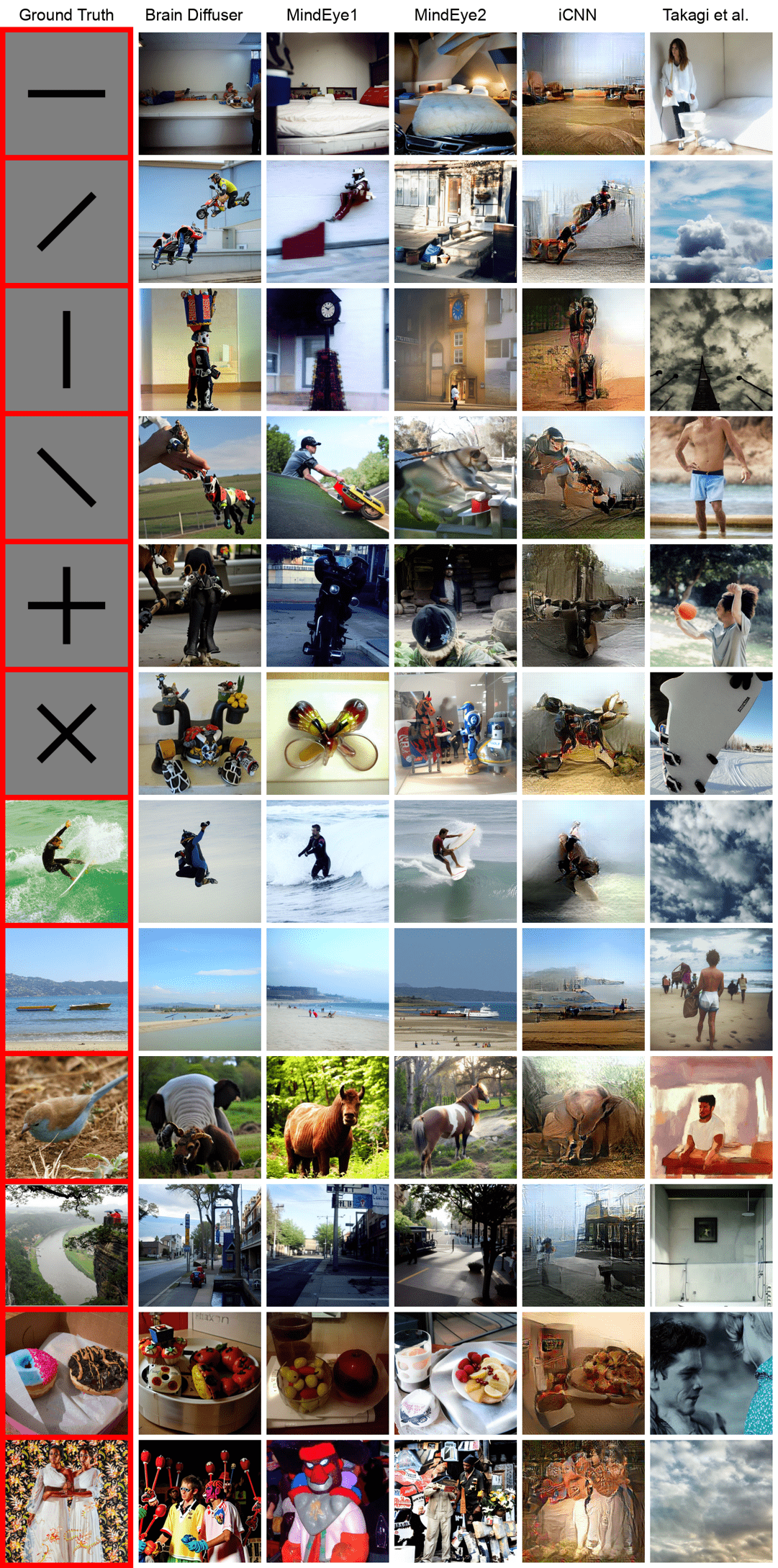}
\caption{Qualitative comparison of the median-case reconstructions on stimuli seen during the vision trials of NSD-Imagery. Samples selected are the median scoring according to the reconstruction metrics in Table $1$ of the manuscript.} 
\label{figure:vision_median}
\end{figure}

\begin{figure}
\centering
\includegraphics[height=0.94\textheight]{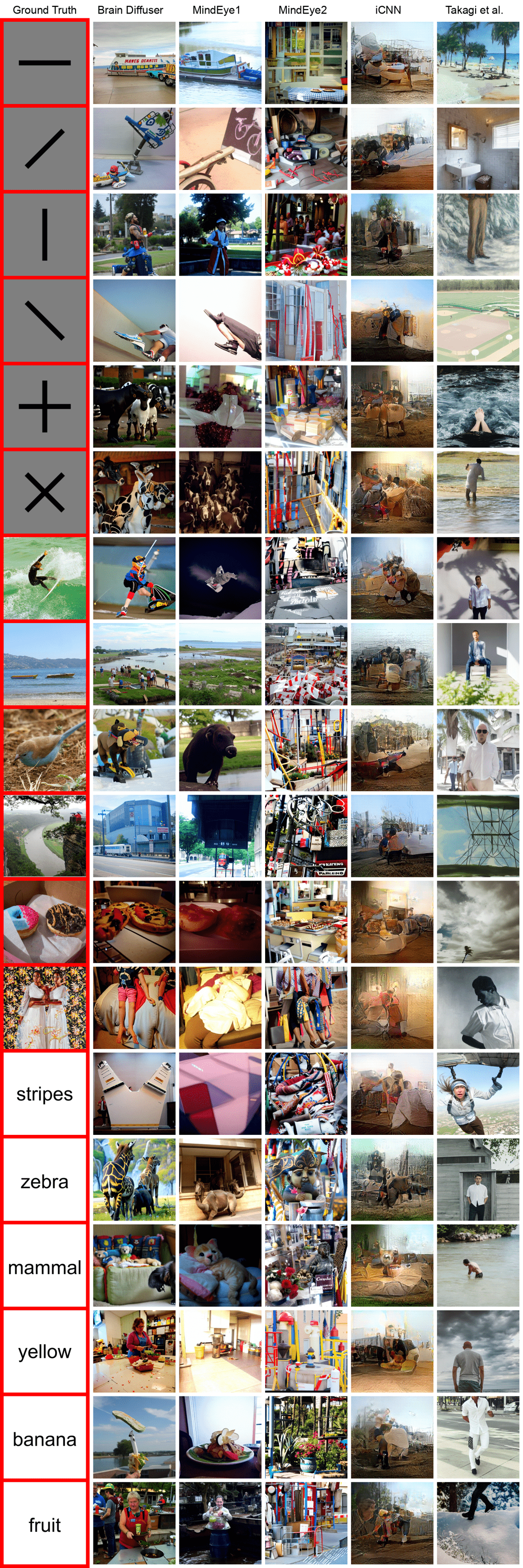}
\caption{Qualitative comparison of the median-case reconstructions on stimuli imagined during the imagery trials of NSD-Imagery. Samples are selected the same way as Figure \ref{figure:vision_median}.} 
\label{figure:imagery_median}
\end{figure}

\begin{figure}[!htb]
\centering
\includegraphics[width=0.87\columnwidth]{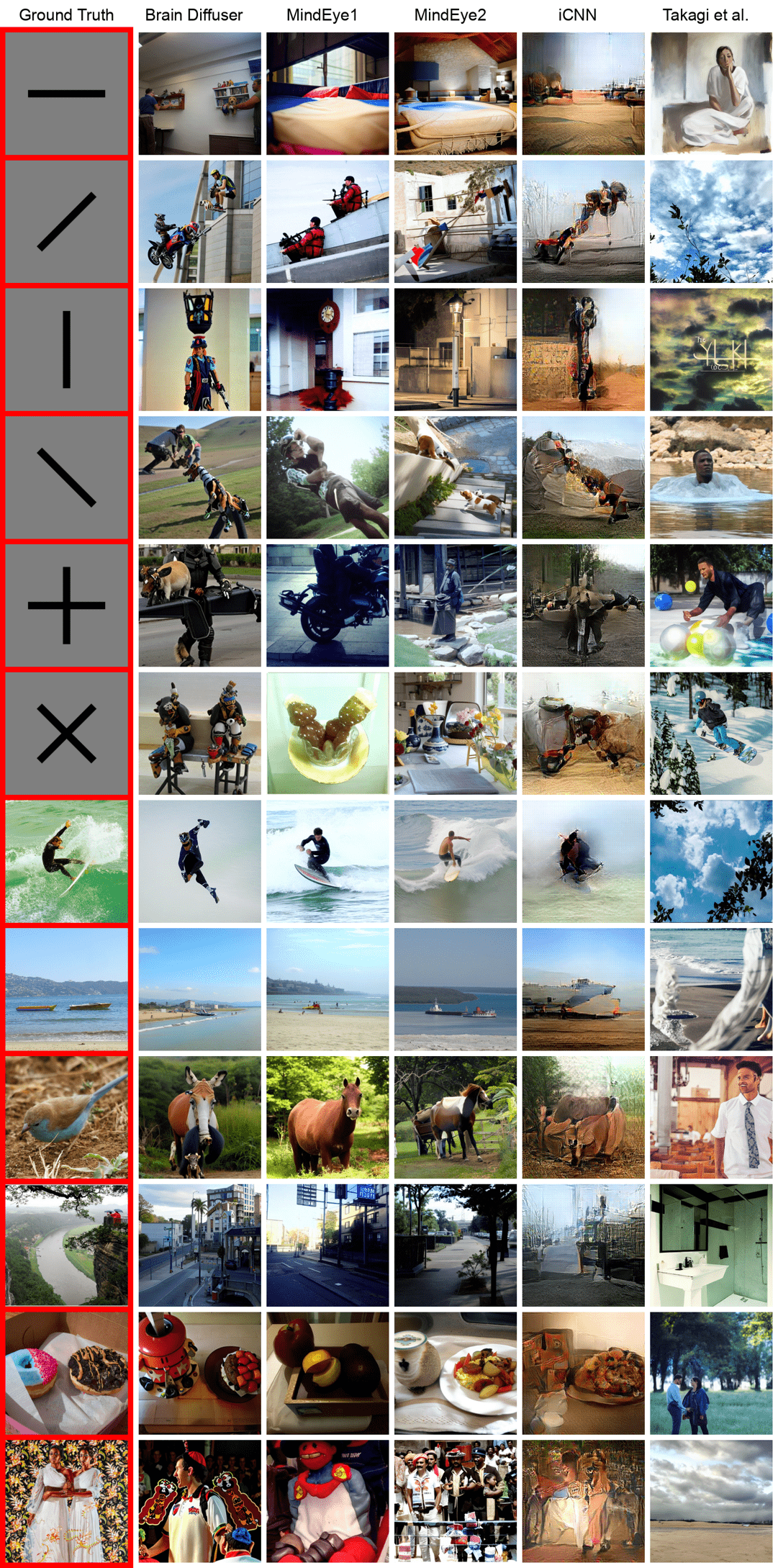}
\caption{Qualitative comparison of the worst-case reconstructions on stimuli seen during the vision trials of NSD-Imagery. Samples selected are the worst scoring according to the reconstruction metrics in Table $1$ of the manuscript.} 
\label{figure:vision_worst}
\end{figure}

\begin{figure}
\centering
\includegraphics[height=0.94\textheight]{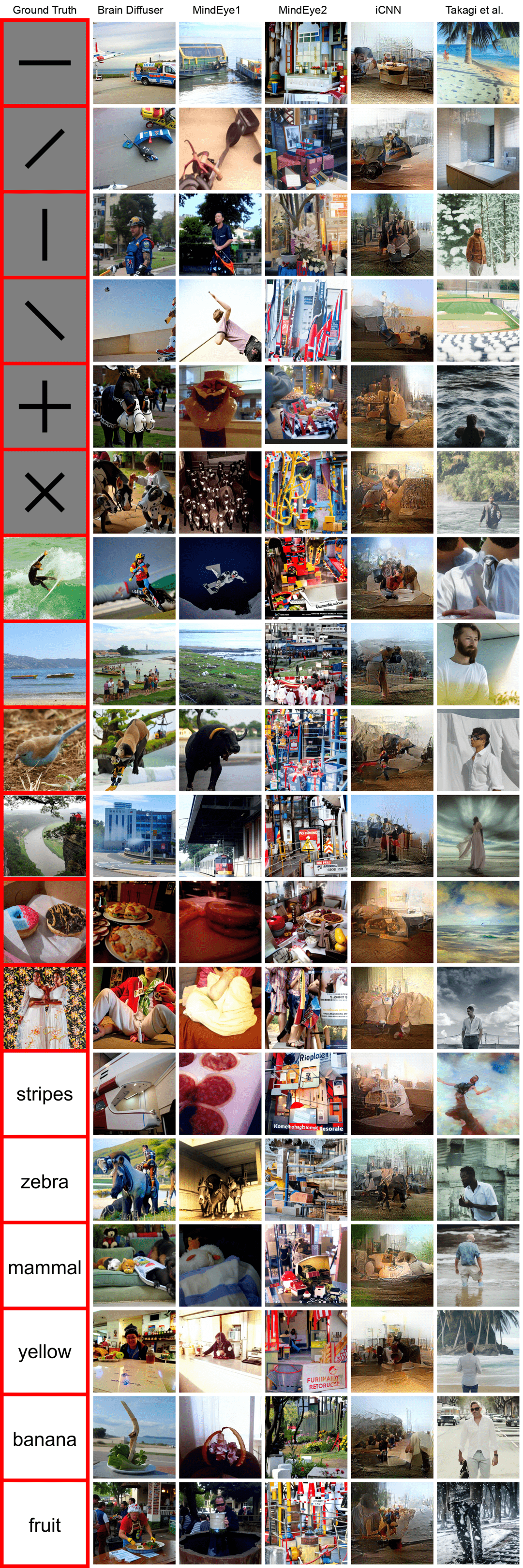}
\caption{Qualitative comparison of the worst-case reconstructions on stimuli imagined during the imagery trials of NSD-Imagery. Samples are selected the same way as Figure \ref{figure:vision_worst}.} 
\label{figure:imagery_worst}
\end{figure}

\FloatBarrier
\onecolumn
\subsection{Comparison of image feature metrics across stimuli types}
\label{app:stimtypes}

\FloatBarrier

\begin{table}[!htb]
    \centering
    \setlength{\tabcolsep}{4pt}
    \small
    \resizebox{\textwidth}{!}{
    \begin{tabular}{lccccccccccc}
        \toprule
        Method & \multicolumn{4}{c}{Low-Level} & \multicolumn{4}{c}{High-Level} & \multicolumn{3}{c}{Brain Correlation} \\
        \cmidrule(lr){2-5} \cmidrule(lr){6-9} \cmidrule(l){10-12}
        & PixCorr $\uparrow$ & SSIM $\uparrow$ & Alex(2) $\uparrow$ & Alex(5) $\uparrow$ & Incep $\uparrow$ & CLIP $\uparrow$  & Eff $\downarrow$ & SwAV $\downarrow$ & Early Vis. $\uparrow$ & Higher Vis. $\uparrow$ & Visual Cortex $\uparrow$ \\
        \midrule
        \multicolumn{12}{c}{\textbf{Mental Imagery Reconstructions (Simple Stimuli)}} \\
        \midrule
        \textbf{MindEye1 \cite{scotti_reconstructing_2023}} & \underline{0.033} & 0.456 & \textbf{43.71\%} & \textbf{61.67\%} & \underline{37.46\%} & \underline{58.37\%} & \textbf{0.974} & \underline{0.563} & \textbf{0.200} & \textbf{0.107} & \textbf{0.148} \\
        \textbf{Brain Diffuser \cite{ozcelik2023braindiffuser}} & 0.013 & \underline{0.524} & \underline{30.68\%} & \underline{50.68\%} & 34.43\% & 44.51\% & 0.983 & 0.603 & \underline{0.152} & \underline{0.091} & \underline{0.128} \\
        \textbf{iCNN \cite{shen_deep_2019}} & \textbf{0.063} & 0.427 & 27.42\% & 47.65\% & \textbf{45.11\%} & \textbf{67.99\%} & 1.006 & \textbf{0.546} & 0.138 & 0.045 & 0.081 \\
        \textbf{MindEye2 \cite{Scotti2024MindEye2}} & 0.011 & 0.448 & 23.37\% & 45.34\% & 31.14\% & 49.02\% & 0.987 & 0.590 & 0.074 & 0.035 & 0.051 \\
        \textbf{Takagi et al. \cite{takagi2022_decoding,takagi2023improving}} & 0.027 &  \textbf{0.595} & 29.70\% & 50.30\% & 37.12\% & 54.70\% & \underline{0.980} & 0.591 & -0.002 & -0.001 & 0.002 \\
        \midrule
        \multicolumn{12}{c}{\textbf{Vision Reconstructions (Simple Stimuli)}} \\
        \midrule
        \textbf{MindEye1 \cite{scotti_reconstructing_2023}} & \underline{0.129} & 0.506 & \textbf{62.01\%} & \textbf{76.36\%} & \underline{43.33\%} &  \underline{60.64\%} & \textbf{0.961} & \underline{0.549} & \underline{0.370} & \textbf{0.140} & \underline{0.243} \\
        \textbf{Brain Diffuser \cite{ozcelik2023braindiffuser}} & 0.075 & \textbf{0.586} & 40.19\% & 66.67\% & 38.30\% & 42.20\% & 0.988 & 0.601 & 0.209 & 0.106 & 0.169 \\
        \textbf{iCNN \cite{shen_deep_2019}} & \textbf{0.132} & 0.454 & \underline{57.01\%} & \underline{74.89\%} & 37.69\% & \textbf{69.02\%} & 0.992 & \textbf{0.534} & \textbf{0.447} & \underline{0.133} & \textbf{0.278} \\
        \textbf{MindEye2 \cite{Scotti2024MindEye2}} & 0.040 & 0.487 & 50.87\% & 68.98\% & \textbf{43.52\%} & 52.46\% & 0.980 & 0.577 & 0.334 & 0.108 & 0.204 \\
        \textbf{Takagi et al. \cite{takagi2022_decoding,takagi2023improving}} & 0.015 & \underline{0.542} & 22.16\% & 50.68\% & 32.73\% & 55.19\% & \underline{0.968} & 0.588 & 0.012 & -0.007 & 0.001 \\
        \bottomrule
    \end{tabular}
    }
    \caption{Quantitative comparison between reconstruction methods for both imagery and vision trials on simple stimuli. Metrics are the same as Table $1$ of the manuscript.}
    \label{table:simple_stimuli}
\end{table}

\begin{table}[!htb]
    \centering
    \setlength{\tabcolsep}{4pt}
    \small
    \resizebox{\textwidth}{!}{
    \begin{tabular}{lccccccccccc}
        \toprule
        Method & \multicolumn{4}{c}{Low-Level} & \multicolumn{4}{c}{High-Level} & \multicolumn{3}{c}{Brain Correlation} \\
        \cmidrule(lr){2-5} \cmidrule(lr){6-9} \cmidrule(l){10-12}
        & PixCorr $\uparrow$ & SSIM $\uparrow$ & Alex(2) $\uparrow$ & Alex(5) $\uparrow$ & Incep $\uparrow$ & CLIP $\uparrow$  & Eff $\downarrow$ & SwAV $\downarrow$ & Early Vis. $\uparrow$ & Higher Vis. $\uparrow$ & Visual Cortex $\uparrow$ \\
        \midrule
        \multicolumn{12}{c}{\textbf{Mental Imagery Reconstructions (Complex Stimuli)}} \\
        \midrule
        \textbf{MindEye1 \cite{scotti_reconstructing_2023}} & \underline{0.138} & 0.243 & \textbf{75.42\%} & 60.34\% & \underline{66.51\%} & 51.06\% & \underline{0.921} & \textbf{0.566} & \textbf{0.159} & \textbf{0.164} & \textbf{0.161} \\
        \textbf{Brain Diffuser \cite{ozcelik2023braindiffuser}} & 0.114 & \underline{0.278} & \underline{73.60\%} & \textbf{66.02\%} & \textbf{71.02\%} & \textbf{63.64\%} & \textbf{0.888} & \underline{0.567} & \underline{0.114} & \underline{0.163} & \underline{0.154} \\
        \textbf{iCNN \cite{shen_deep_2019}} & \textbf{0.153} & 0.253 & 73.71\% & 62.84\% & 53.67\% & 15.46\% & 0.982 & 0.575 & 0.089 & 0.079 & 0.081 \\
        \textbf{MindEye2 \cite{Scotti2024MindEye2}} & 0.032 & 0.231 & 70.42\% & \underline{65.11\%} & 61.97\% & \underline{51.93\%} & 0.943 & 0.601 & 0.062 & 0.074 & 0.068 \\
        \textbf{Takagi et al. \cite{takagi2022_decoding,takagi2023improving}} & -0.039 & \textbf{0.315} & 54.05\% & 30.08\% & 49.39\% & 25.46\% & 0.972 & 0.622 & -0.001 & 0.009 & 0.003 \\
        \midrule
        \multicolumn{12}{c}{\textbf{Vision Reconstructions (Complex Stimuli)}} \\
        \midrule
        \textbf{MindEye1 \cite{scotti_reconstructing_2023}} & \underline{0.308} & \underline{0.318} & \underline{85.11\%} & \underline{85.27\%} & 81.55\% & 70.04\% & \underline{0.800} & \underline{0.471} & \underline{0.378} & \textbf{0.365} & \underline{0.379} \\
        \textbf{Brain Diffuser \cite{ozcelik2023braindiffuser}} & 0.139 & \textbf{0.323} & 80.49\% & 79.02\% & \underline{83.60\%} & \underline{74.43\%} & 0.829 & 0.509 & 0.284 & 0.353 & 0.341 \\
        \textbf{iCNN \cite{shen_deep_2019}} & \textbf{0.316} & 0.316 & \textbf{86.33\%} & \textbf{87.80\%} & \textbf{84.62\%} & 29.05\% & 0.860 & 0.514 & \textbf{0.437} & 0.358 & \textbf{0.397} \\
        \textbf{MindEye2 \cite{Scotti2024MindEye2}} & 0.223 & 0.333 & 84.28\% & 85.83\% & 80.08\% & \textbf{77.46\%} & \textbf{0.794} & \textbf{0.454} & 0.378 & \underline{0.360} & 0.376 \\
        \textbf{Takagi et al. \cite{takagi2022_decoding,takagi2023improving}} & -0.041 & 0.281 & 60.95\% & 27.84\% & 45.80\% & 30.83\% & 0.971 & 0.632 & -0.024 & 0.039 & 0.016 \\
        \bottomrule
    \end{tabular}
    }
    \caption{Quantitative comparison between reconstruction methods for both imagery and vision trials on complex stimuli. Metrics are the same as Table $1$ of the manuscript.}
    \label{table:complex_stimuli}
\end{table}

\FloatBarrier
\twocolumn

\subsection{Behavioral experiment}
\label{app:behavioral}

\subsubsection{Experiment protocols}
\label{app:experimentprotocols}
We conducted a set of behavioral experiments on $500$ human raters online. For our experiment, we identified no risks to the human participants, and our institution's IRB approved our experiment. We probed $3$ experiments intermixed into two discrete sections within the same behavioral tasks, with each experiment consisting of trials sampled evenly from the different stimulus types and the 4 NSD subjects who completed all 40 scanning sessions (subjects 1, 2, 5, 7). The experimental trials within each task were shuffled and $36$ trials were presented to each subject. Our subjects were recruited through the \href{http://www.prolific.ac.uk)}{Prolific platform}, with our experimental tasks hosted on \href{http://meadows-research.com}{Meadows}. Each human rater was paid $\$1.50$ for the completion of the experiment, and the median completion time was $6$ minutes and $17$ seconds, resulting in an average payment rate of $\$14.32$/hour. Each human rater was presented with $6$ attention check trials during the experiment. An attention check is a trial in which the ground truth image is presented as a candidate image during the trial. Because the ground truth image will always be the image that is most similar to itself, these trials were used to identify whether subjects were paying attention to the task and the instructions. We identified $5$ human raters who failed at least 2 attention checks and removed those raters from our data before conducting our analysis. Code to reproduce our experiment can be found in \href{https://anonymous.4open.science/r/mental_imagery_behavioral_analysis-2214/}{our anonymized GitHub repository.}

\subsubsection{2AFC identification task}
\begin{figure}[!htb]
\setlength{\abovecaptionskip}{-10pt}    
\setlength{\belowcaptionskip}{-10pt}    
\begin{center}
\includegraphics[width=\columnwidth]{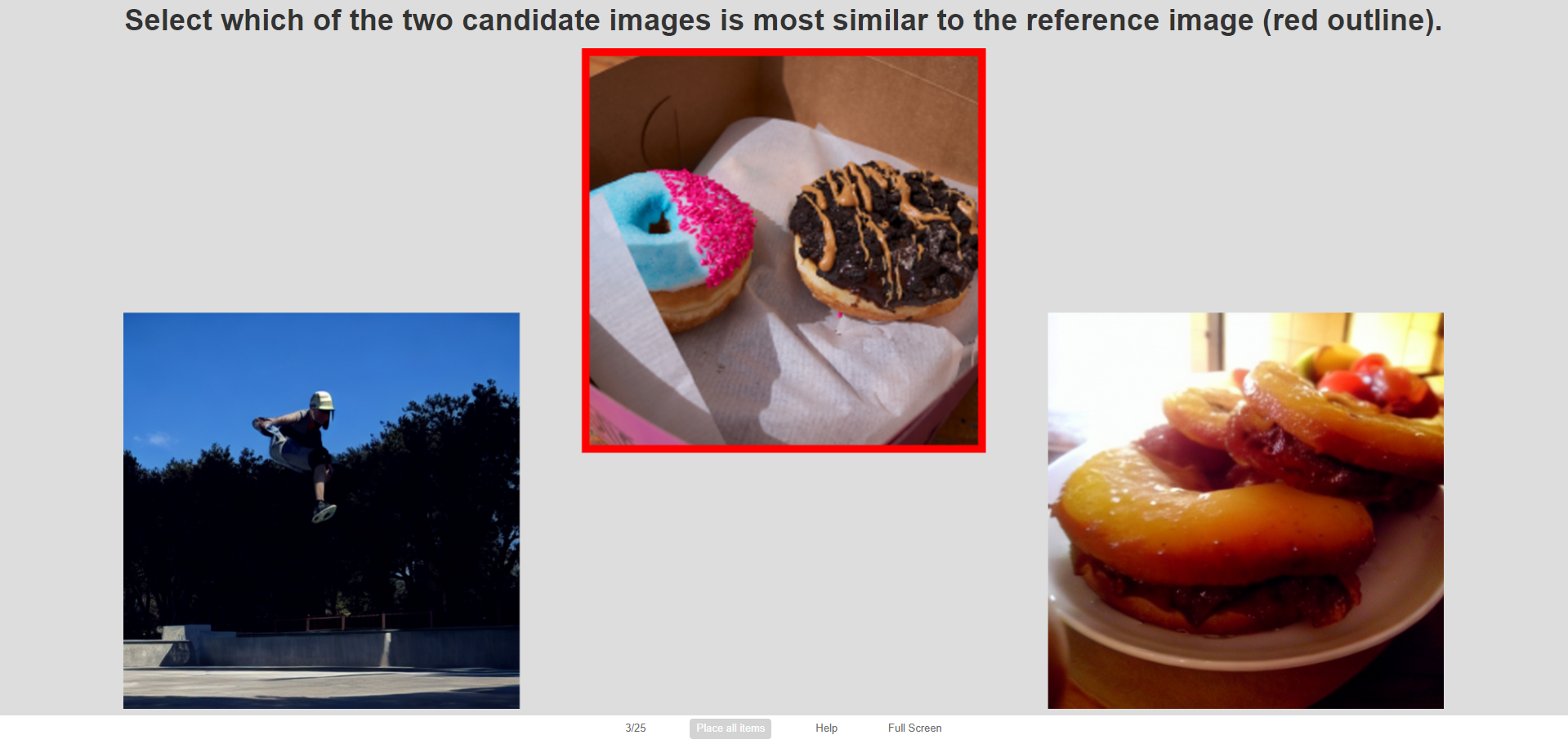}
\end{center}
\caption{An example of the 2 alternative forced choice task used in the first behavioral experiment performed by human raters.} 
\label{figure:task1}
\end{figure}

Our first experiment was a $2$ alternative forced choice task (2AFC) facilitated by the "Match-To-Sample" task on the Meadows platform. An example of the first experiment can be seen in Figure \ref{figure:task1}. In this experiment, human raters were asked to select which of two candidate images was more similar to a reference image. The reference image provided is the ground truth image the NSD-Imagery subject either saw or imagined, and the $2$ candidate images were the target reconstruction of the reference image, or a randomly selected reconstruction from an fMRI scan corresponding to a different stimulus of the same stimulus type. The two candidate images were always sampled from the same reconstruction method and NSD-Imagery subject. This experiment was repeated for all reconstruction methods, visual modalities, NSD subjects, and across 10 reconstructions sampled from the output distribution of each reconstruction method. With the results presented in Section $4.5$, we establish a baseline for human-rated image identification accuracy of mental image reconstructions, as no other paper has conducted behavioral evaluations of mental image reconstructions.

\subsubsection{Continuous similarity rating task}
\begin{figure}[!htb]
\setlength{\abovecaptionskip}{-10pt}    
\setlength{\belowcaptionskip}{-10pt}    
\begin{center}
\includegraphics[width=\columnwidth]{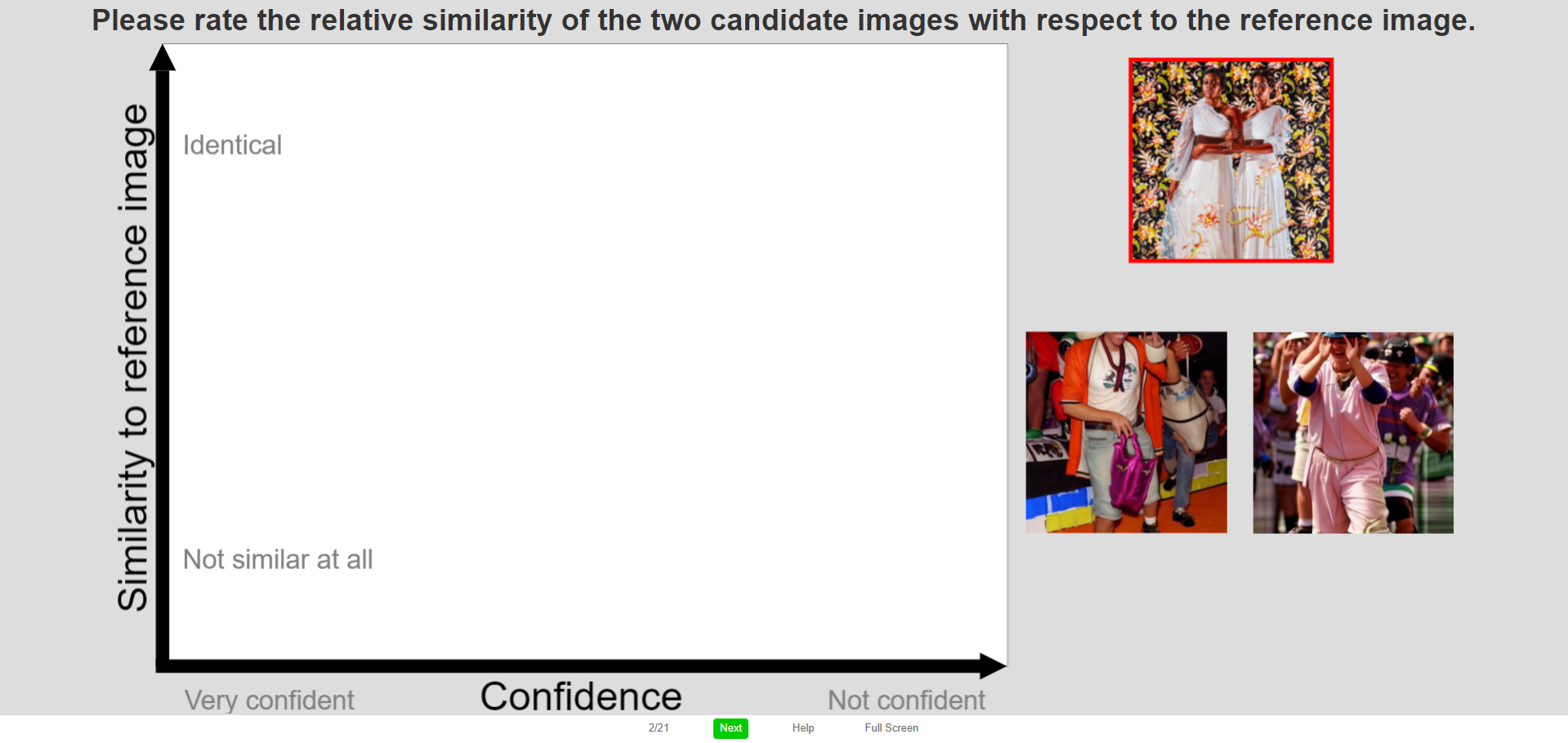}
\end{center}
\caption{An example of similarity score task used in experiment $2$ of the behavioral experiment performed by human raters.} 
\label{figure:task23}
\end{figure}

The second experiment we conducted was facilitated by the "Drag-Rate" task on the Meadows platform. An example of the task can be seen in Figure \ref{figure:task23}. In this task, human raters were presented with a reference image, two candidate images, and a continuous two-dimensional plot that they could drag the candidate images onto, where the Y-axis represented "similarity to the reference image" and the X-axis represented the rater's confidence. The reference image provided was always the ground truth image the NSD-Imagery subject either saw or imagined. For experiment $2$, the $2$ candidate images were reconstructions of the reference image from the imagery and vision trials of the NSD-Imagery trials. Experiment $2$ was repeated for the simple and complex stimuli (as conceptual stimuli do not have meaningful vision reconstructions), all reconstruction methods, NSD subjects, and across 10 reconstructions sampled from the output distribution of each reconstruction method. One-dimensional similarity ratings—like the ones used in this section of the experiment—can be extremely sensitive to the context of the alternative samples being compared against, and so are primarily useful for comparing the relative similarity of the candidate stimuli presented during each individual trial. The task was designed with this in mind, configured to directly compare the difference in quality between reconstructions of vision and imagery for each method. Our analysis of these results in Section $4.5$ provides a detailed analysis of how reconstruction performance scales across vision and imagery.

\subsection{iCCN implementation}
\label{app:icnnchanges}
Originally introduced in \citet{shen_deep_2019}, and first trained on NSD in  \citet{Shirakawa2024SpuriousRF}, we adapt the author's open source implementation to try and faithfully replicate their results, making the following changes to the implementation:
\begin{enumerate}
    \item \textbf{Normalization of images:} We disabled normalization of images when computing VGG19 features. During our initial trials, normalization led to unexpected color distortions in the reconstructed images. Removing normalization allowed the reconstructions to maintain their original color integrity, which is particularly crucial for visual comparisons in tasks requiring precise color representation.
    \item \textbf{Feature decoding with Ridge Regression:} Instead of the \texttt{fastl2lir} library, we employed the Ridge Regression implementation from the \texttt{sklearn} library. This change enhanced compatibility with the rest of our workflow and provided better support for managing memory-intensive computations. For VGG19 layers with a large feature space, feature decoding was performed in chunks. This approach enabled the simultaneous calculation of features and fitting of the Ridge Regression model without requiring intermediate results to be saved to disk, thereby optimizing both time and memory usage.
\end{enumerate}

\subsection{Impact of trial repetition averaging on performance}
\label{app:trial_reps}

\begin{figure}[!htb]
\vspace{-15pt}
\setlength{\abovecaptionskip}{2pt}    
\setlength{\belowcaptionskip}{-5pt}    
\centering
\includegraphics[width=\columnwidth]{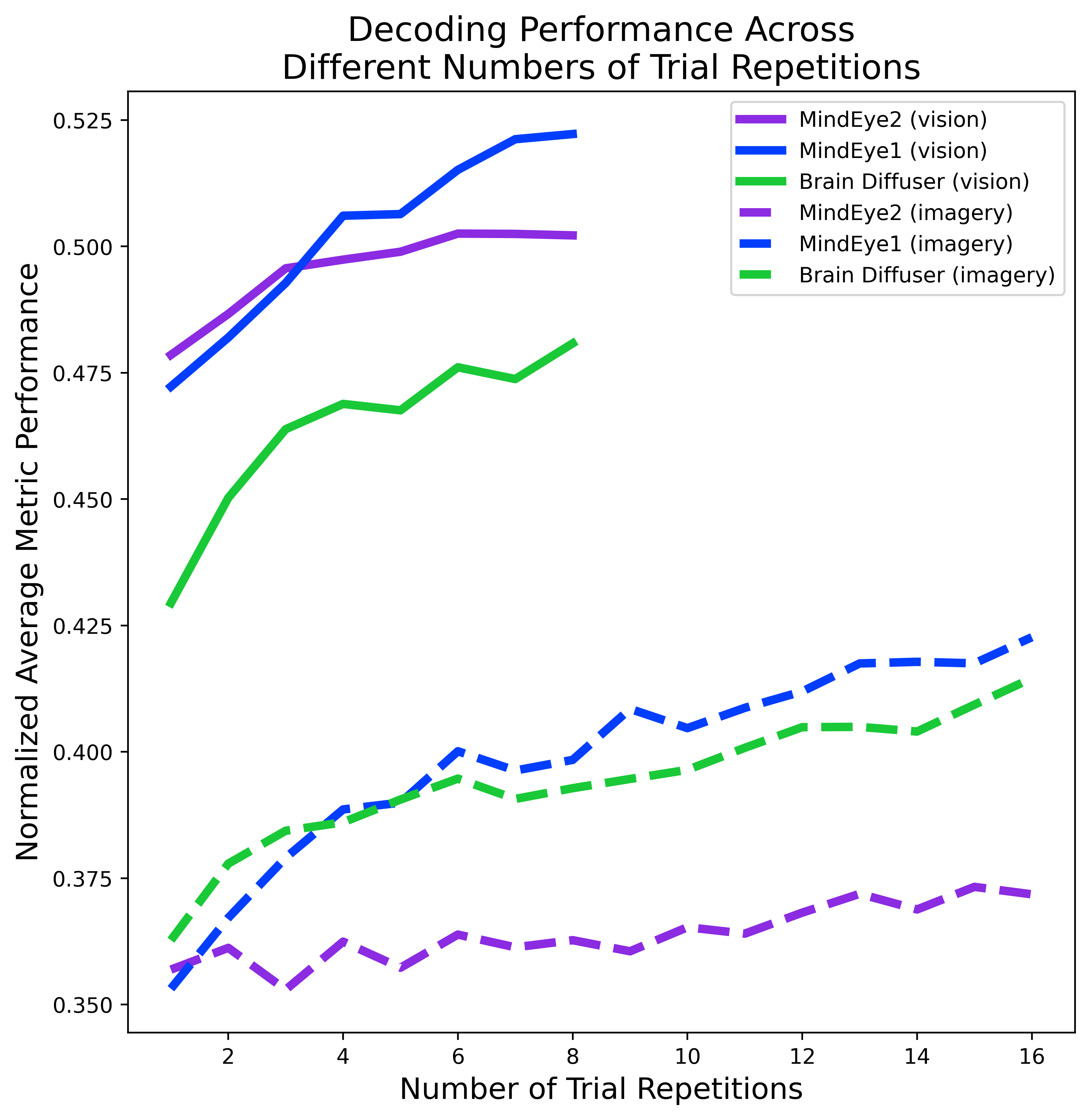}
\caption{Performance of various methods when averaging across brain activity responses to multiple trial repetitions of the same stimulus. Y axis is the normalized average of all metrics in Table $1$ of the manuscript, X axis is the number of averaged trial repetitions.} 
\label{figure:trial_reps}
\end{figure}

One of the experimental details that varies between NSD \cite{allen_massive_2022} and NSD-Imagery is the number of times each stimulus was presented in the experiment, also called the number of trial repetitions. NSD contained $3$ trial repetitions of each stimulus in both the training and test sets, while NSD-Imagery contains 8 trial repetitions for the vision task and $16$ trial repetitions for the imagery task. In Figure \ref{figure:trial_reps}, we plot the effect of these additional trial repetitions on the performance of a subset of the reconstruction methods evaluated in this work.

\subsection{Impact of training data scale on performance}
\label{app:scaling}

\begin{figure}[!htb]
\vspace{-5pt}
\setlength{\abovecaptionskip}{2pt}    
\setlength{\belowcaptionskip}{-5pt}    
\centering
\includegraphics[width=\columnwidth]{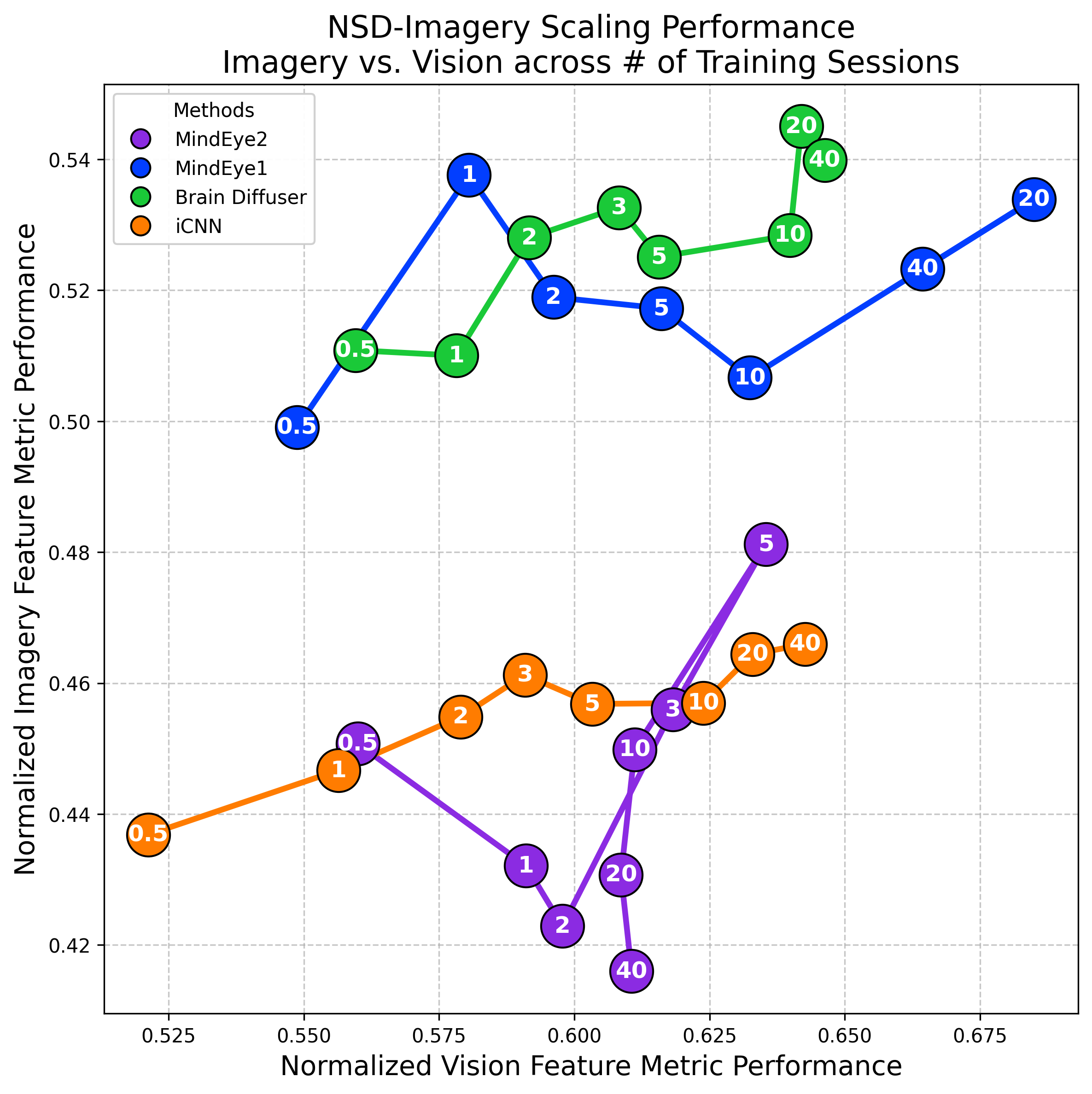}
\caption{Performance of various methods on NSD-Imagery for Subject 1 when trained on different numbers of fMRI sessions present in NSD. Each session includes approximately one hour of fMRI data. Metrics are the normalized average of all metrics in Table $1$ of the manuscript, with imagery performance on the Y axis and vision on the X axis. Methods are indicated by color, with the number of training sessions indicated by the numbers in each dot.} 
\label{figure:scaling}
\end{figure}

An additional challenge in deploying these fMRI-to-image decoding methods lies in making them more generalizable to new subjects. 
All of the methods examined in this paper were trained with $40$ hours of subject-specific fMRI data comprising $10,000$ unique stimuli. Collecting this much training data for new subjects in clinical settings is currently impractical or impossible for certain patients.
Recent work in MindEye2 \cite{Scotti2024MindEye2} has made strides in scaling decoding procedures using a multi-subject pretraining step, however as demonstrated in Figure \ref{figure:scaling}, this approach generalizes poorly to mental imagery data. We additionally note that the methods that used ridge regression decoding backbones (Brain Diffuser, iCNN) produce much more consistent scaling improvements on mental images than the models that utilize deep neural network backbones (MindEye1, MindEye2).